\documentclass{article}

\usepackage[table,xcdraw]{xcolor}
\definecolor{lightgrey}{HTML}{E7E7E7}

\usepackage{microtype}
\usepackage{graphicx}
\usepackage{subfigure}
\usepackage{booktabs} % for professional tables
\usepackage{multirow}

\usepackage{hyperref}

\usepackage[accepted]{icml2025}

\usepackage{amsmath}
\usepackage{amssymb}
\usepackage{mathtools}
\usepackage{amsthm}

\usepackage[capitalize,noabbrev]{cleveref}

\theoremstyle{plain}

\theoremstyle{definition}

\theoremstyle{remark}

\usepackage[textsize=tiny]{todonotes}

\usepackage{xspace}
\newcommand{\hipporag}{HippoRAG\xspace}
\newcommand{\ours}{HippoRAG 2\xspace}

\icmltitlerunning{From RAG to Memory: Non-Parametric Continual Learning for Large Language Models}

\begin{document}

\twocolumn[
\icmltitle{From RAG to Memory: Non-Parametric Continual Learning for \\ Large Language Models}



\icmlsetsymbol{equal}{*}

\begin{icmlauthorlist}
\icmlauthor{Bernal Jim\'{e}nez Guti\'{e}rrez}{equal,yyy}
\icmlauthor{Yiheng Shu}{equal,yyy}
\icmlauthor{Weijian Qi}{yyy}
\icmlauthor{Sizhe Zhou}{zzz}
\icmlauthor{Yu Su}{yyy}
\end{icmlauthorlist}

\icmlaffiliation{yyy}{The Ohio State University, Columbus, OH, USA}
\icmlaffiliation{zzz}{University of Illinois Urbana-Champaign, IL, USA}

\icmlcorrespondingauthor{Bernal Jim\'{e}nez Guti\'{e}rrez}{jimenezgutierrez.1@osu.edu}
\icmlcorrespondingauthor{Yiheng Shu}{shu.251@osu.edu} 
\icmlcorrespondingauthor{Yu Su}{su.809@osu.edu} 

\icmlkeywords{Machine Learning, ICML}

\vskip 0.3in
]



\printAffiliationsAndNotice{\icmlEqualContribution} % otherwise use the standard text.

\begin{abstract}

Our ability to continuously acquire, organize, and leverage knowledge is a key feature of human intelligence that AI systems must approximate to unlock their full potential. 
Given the challenges in continual learning with large language models (LLMs), retrieval-augmented generation (RAG) has become the dominant way to introduce new information. 
However, its reliance on vector retrieval hinders its ability to mimic the dynamic and interconnected nature of human long-term memory. 
Recent RAG approaches augment vector embeddings with various structures like knowledge graphs to address some of these gaps, namely sense-making and associativity.
However, their performance on more basic factual memory tasks drops considerably below standard RAG.
We address this unintended deterioration and propose \ours, a framework that outperforms standard RAG comprehensively on factual, sense-making, and associative memory tasks. 
\ours builds upon the Personalized PageRank algorithm used in HippoRAG and enhances it with deeper passage integration and more effective online use of an LLM. 
This combination pushes this RAG system closer to the effectiveness of human long-term memory, achieving a 7\% improvement in associative memory tasks over the state-of-the-art embedding model while also exhibiting superior factual knowledge and sense-making memory capabilities.
This work paves the way for non-parametric continual learning for LLMs. 
Code and data are available at \url{https://github.com/OSU-NLP-Group/HippoRAG}.

\end{abstract}

\section{Introduction}

\begin{figure*}
    \centering
    \includegraphics[width=\linewidth]{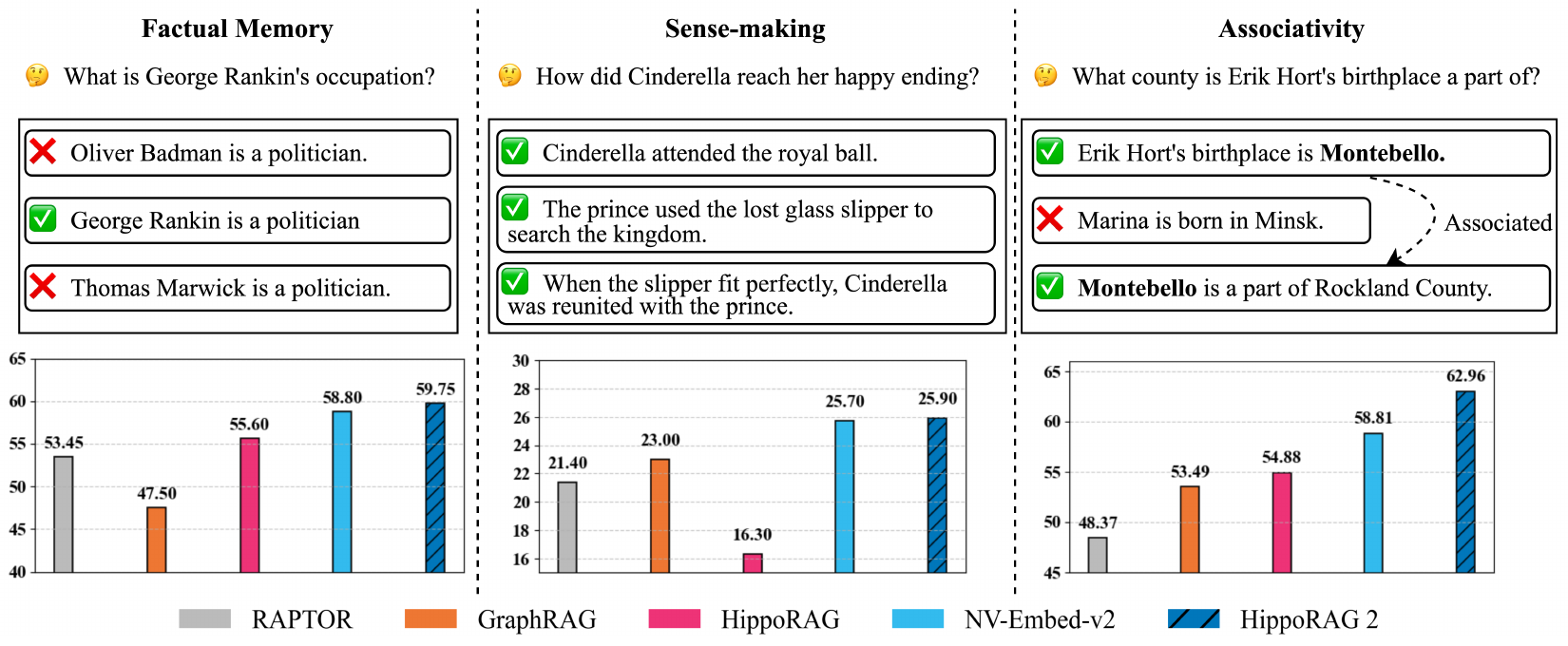}
  \vskip -0.1in
    \caption{
    Evaluation of continual learning capabilities across three key dimensions: factual memory (NaturalQuestions, PopQA), sense-making (NarrativeQA), and associativity (MuSiQue, 2Wiki, HotpotQA, and LV-Eval). 
    \ours surpasses other methods across all benchmark categories, bringing it one step closer to a true long-term memory system.
    }
    \label{fig:introduction}
\end{figure*}

In an ever-evolving world, the ability to continuously absorb, integrate, and leverage knowledge is one of the most important features of human intelligence.
From lawyers navigating shifting legal frameworks to researchers tracking multifaceted scientific progress, much of our productivity relies on this incredible capacity for continual learning.
It is imperative for AI systems to approximate this capability in order to become truly useful human-level assistants.
% -long a central pursuit of the continual learning community.

In recent years, large language models (LLMs) have made remarkable progress in many aspects of human intelligence.
However, efforts to endow these models with our evolving long-term memory capabilities have faced significant challenges in both fully absorbing new knowledge \cite{zhong-etal-2023-mquake, hoelscher-obermaier-etal-2023-detecting} and avoiding catastrophic forgetting \cite{cohen-etal-2024-evaluating, gu-etal-2024-model}, due to the complex distributional nature of their parametric knowledge.
Retrieval-augmented generation (RAG) has emerged as a way to circumvent these obstacles and allow LLMs to access new information in a \textit{non-parametric} fashion without altering an LLM's parametric representation.
Due to their simplicity and robustness \cite{zhong-etal-2023-mquake, xie2024adaptive}, RAG has quickly become the \textit{de facto} continual learning solution for production LLM systems.
However, their reliance on simple vector retrieval results in the inability to capture two vital aspects of our interconnected long-term memory system: \textbf{sense-making} (\citet{klein2006making}; the ability to interpret larger, more complex, or uncertain contexts) and \textbf{associativity} (\citet{suzuki2005associative}; the capacity to draw multi-hop connections between disparate pieces of knowledge).

Several RAG frameworks that engage an LLM to explicitly structure its retrieval corpus have been recently proposed to address these limitations. 
To enhance sense-making, such \textit{structure-augmented} RAG methods allow an LLM to either generate summaries \cite{graphrag, raptor, chen2023walkingmemorymazecontext} or a knowledge graph (KG) structure \cite{lightrag} to link groups of disparate but related passages, thereby improving the RAG system's ability to understand longer and more complex discourse such as long stories.
To address the associativity gap, the authors of HippoRAG \cite{hipporag} use the Personalized PageRank algorithm \cite{david02topic} and an LLM's ability to automatically construct a KG and endow the retrieval process with multi-hop reasoning capabilities.

Although these methods demonstrate strong performance in both of these more challenging memory tasks, bringing RAG truly closer to human long-term memory requires robustness across simpler memory tasks as well.
% Although each of these methods demonstrate strong performance on one of these challenging memory tasks or the other, bringing RAG truly closer to human long-term memory requires a system that can not only mimic the strengths of standard RAG but also achieves robust performance on both of these complex tasks.
In order to understand whether these systems could achieve such robustness, we conduct comprehensive experiments that not only simultaneously evaluate their associativity and sense-making capacity through multi-hop QA and large-scale discourse understanding, but also test their \textbf{factual memory} abilities via simple QA tasks, which standard RAG is already well-equipped to handle.

As shown in Figure \ref{fig:introduction}, our evaluation reveals that all previous structure-augmented methods underperform against the strongest embedding-based RAG methods available on all three benchmark types.
Perhaps unsurprisingly, we find that each method type experiences the largest performance decay in tasks outside its own experimental setup. For example, HippoRAG's performance drops most on large-scale discourse understanding due to its lack of query-based contextualization, while RAPTOR's performance deteriorates substantially on the simple and multi-hop QA tasks due to the noise introduced into the retrieval corpora by its LLM summarization mechanism.

In this work, we leverage this experimental setting to help us address the robustness limitations of these innovative approaches while avoiding the pitfalls of focusing too narrowly on just one task.
Our proposed method, \ours, leverages the strength of HippoRAG's OpenIE and Personalized PageRank (PPR) methodologies while addressing its query-based contextualization limitations by integrating passages into the PPR graph search process, involving queries more deeply in the selection of KG triples as well as engaging an LLM in the online retrieval process to recognize when retrieved triples are irrelevant.

Through extensive experiments, we find that this design provides \ours with consistent performance improvements over the most powerful standard RAG methods across the board.
More specifically, our approach achieves an average $7$ point improvement over standard RAG in associativity tasks while showing no deterioration and even slight improvements in factual memory and sense-making tasks. 
Furthermore, we show that our method is robust to different retrievers as well as to the use of strong open-source and proprietary LLMs, allowing for a wide degree of usage flexibility.
All of these results suggest that \ours is a promising step in the development of a more human-like non-parametric continual learning system for LLMs.

\section{Related Work}

\subsection{Continual Learning for LLMs}

As the use of LLMs in real-world applications grows, it becomes increasingly important for them to acquire and integrate new knowledge over time while preserving past information---as evidenced by the many benchmarking efforts in this direction \cite{zhong-etal-2023-mquake, streamingqa, kim-etal-2024-carpe, cont_pretrain_roth_multimodal, ticlm}. Given the high computational cost of full-scale LLM pretraining, various techniques have been leveraged to endow these models with this continual learning capacity. These approaches generally fall into three categories: continual fine-tuning, model editing, and RAG \cite{shi2024continual}.

\textbf{Continual fine-tuning} involves periodically training an LLM on new data. This can be achieved through methods like continual pretraining \cite{lifelong}, instruction tuning \cite{citb}, and alignment fine-tuning \cite{copr}. 
While effective in incorporating new linguistic patterns and reasoning skills, continual fine-tuning suffers from catastrophic forgetting \cite{huang24mitigating}, where previously learned knowledge is lost as new data is introduced. Moreover, its computational expense makes frequent updates impractical for real-world applications.

\textbf{Model editing} techniques \cite{yao23editing} provide a more lightweight alternative by directly modifying specific parameters in the model to update its knowledge. However, these updates have been found to be highly localized, having little effect on information associated with the update that should also be changed.

\textbf{RAG} has emerged as a scalable and practical alternative for continual learning. Instead of modifying the LLM itself, RAG retrieves relevant external information at inference time, allowing for real-time adaptation to new knowledge. 
We will discuss several aspects of this non-parametric continual learning solution for LLMs in the next section.

\begin{figure*}
    \centering
    \includegraphics[width=\linewidth]{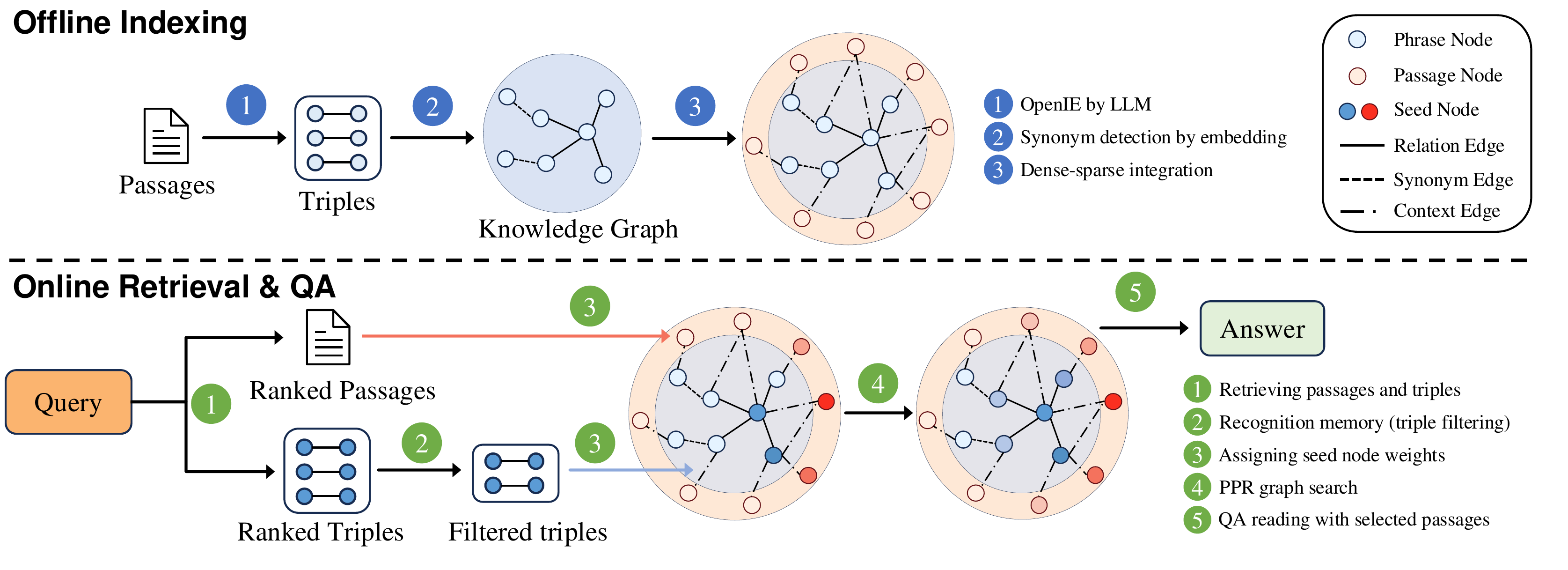}
      \vskip -0.1in
    \caption{\textbf{\ours methodology}.
    For offline indexing, we use an LLM to extract open KG triples from passages, with synonym detection applied to phrase nodes. Together, these phrases and passages form the open KG. 
    For online retrieval, an embedding model scores both the passages and triples to identify the seed nodes of both types for the Personalized PageRank (PPR) algorithm. Recognition memory filters the top triples using an LLM. The PPR algorithm then performs context-based retrieval on the KG to provide the most relevant passages for the final QA task. The different colors shown in the KG nodes above reflect their probability mass; darker shades indicate higher probabilities induced by the PPR process.
    }
    \label{fig:methodology}
  \vskip -0.1in
\end{figure*}

\subsection{Non-Parametric Continual Learning for LLMs}

% \textbf{Encoder model improvements}, particularly through the use of LLM backbones, have led to significant improvements in RAG systems.
% % Recent advancements in embedding models, particularly those utilizing LLMs, have significantly improved RAG systems. 
% By generating high-quality text embeddings, well-trained LLMs can more accurately capture semantic relationships between queries and documents, leading to enhanced retrieval precision and recall. This improvement allows RAG methods to better incorporate relevant external information, enhancing generation accuracy.
% Recent embedding models \cite{gte,gritlm,nvembedv2} often employ LLMs, extensive corpora, improved architectural designs, and instruction fine-tuning techniques, resulting in substantial retrieval performance gains. NV-Embed-v2 \cite{nvembedv2}, as a representative model, serves as the primary comparison method in this paper.
\textbf{Encoder model improvements}, particularly with LLM backbones, have significantly enhanced RAG systems by generating high-quality embeddings that better capture semantic relationships, improving retrieval quality for LLM generation. Recent models \cite{gte,gritlm,nvembedv2} leverage LLMs, large corpora, improved architectures, and instruction fine-tuning for notable retrieval gains. NV-Embed-v2 \cite{nvembedv2} serves as the primary comparison in this paper.

\noindent \textbf{Sense-making} is the ability to understand large-scale 
or complex events, experiences, or data \cite{koli-etal-2024-sensemaking}. Standard RAG methods are limited in this capacity since they require integrating information from disparate passages, and thus, several RAG frameworks have been proposed to address it. RAPTOR \cite{raptor} and GraphRAG \cite{graphrag} both generate summaries that integrate their retrieval corpora. However, they follow distinct processes for detecting what to summarize and at what granularity. While RAPTOR uses a Gaussian Mixture Model to detect document clusters to summarize, GraphRAG uses a graph community detection algorithm that can summarize documents, entity clusters with relations, or a combination of these elements.
LightRAG \cite{lightrag} employs a dual-level retrieval mechanism to enhance comprehensive information retrieval capabilities in both low-level and high-level knowledge, integrating graph structures with vector retrieval. 

Although both GraphRAG and LightRAG use a KG just like our \ours approach, our KG is used to aid in the retrieval process rather than to expand the retrieval corpus itself. This allows \ours to introduce less LLM-generated noise, which deteriorates the performance of these methods in single and multi-hop QA tasks.

\noindent \textbf{Associativity} is the capacity to draw multi-hop connections between disparate facts for efficient retrieval. It is an important part of continual learning, which standard RAG cannot emulate due to its reliance on independent vector retrieval. HippoRAG \cite{hipporag} is the only RAG framework that has addressed this property by leveraging the PPR algorithm over an explicitly constructed open KG. \ours is closely inspired by HippoRAG, which allows it to perform very well on multi-hop QA tasks. However, its more comprehensive integration of passages, queries, and triples allows it to have a more comprehensive performance across sense-making and factual memory tasks as well.

\section{\ours}
\label{sec:long-term}

\subsection{Overview}
% Give a figure for the overview method.

HippoRAG \cite{hipporag} is a neurobiologically inspired long-term memory framework for LLMs in which each component is inspired by its neurobiological analog for human memory. The framework consists of three primary components: 1) an LLM that acts as an artificial neocortex, 2) a KG and the Personalized PageRank algorithm to mirror the auto-associative qualities of the hippocampus and 3) a retrieval encoder that links these two components, reflecting one of the functions of the parahippocampal regions. These components collaborate to replicate the interactions observed in human long-term memory.

HippoRAG's offline indexing process uses an LLM to process passages into KG triples, which are then incorporated into the KG, our artificial hippocampal index. Meanwhile, the retrieval encoder is responsible for detecting synonymy to interconnect information.
In HippoRAG's online retrieval process, the LLM neocortex extracts named entities from a query while the retrieval encoder finds their most similar counterparts in the KG. 
Then, the nodes in the KG corresponding to these entities, which we refer to as seed nodes, are used to run the Personalized PageRank (PPR) algorithm.
More specifically, these seed nodes are used to assign the \textbf{reset probabilities} within PPR, which alter the original PageRank algorithm to distribute probability towards the seed nodes and their neighborhoods, enabling HippoRAG's context-based retrieval.
Although HippoRAG seeks to construct memory from non-parametric RAG, its effectiveness is hindered by a critical flaw: an entity-centric approach that causes context loss during both indexing and inference, as well as difficulties in semantic matching.

Built on the neurobiologically inspired long-term memory framework proposed in HippoRAG \cite{hipporag}, the structure of \ours follows a similar two-stage process: offline indexing and online retrieval, as shown in Figure \ref{fig:methodology}. Additionally, however, \ours introduces several key refinements that improve its alignment with human memory mechanisms:
1) It seamlessly integrates conceptual and contextual information within the KG, enhancing the comprehensiveness and atomicity of the constructed index (\S\ref{subsec:index_integration}).
2) It facilitates more context-aware retrieval by leveraging the KG structure beyond isolated KG nodes (\S\ref{subsec:query_to_triple}).
3) It incorporates recognition memory to improve seed node selection for graph search (\S\ref{subsec:recognition_memory}).
In the following sections, we introduce the pipeline in more detail and elaborate on each of these refinements.

\noindent \textbf{Offline Indexing.}
1) \ours, just as HippoRAG, leverages an LLM to extract triples from each passage using OpenIE, which allows the relations and entities to be generated without any constraints or schema. These triples are then arranged into our schema-less KG or hippocampal index.  We call the subject or object of these triples \textbf{phrases} and the edge connecting them \textbf{relation edge}.
2) Next, the retrieval encoder identifies synonyms by evaluating phrase pairs within the KG, detecting those with vector similarity above a predefined threshold, and adding \textbf{synonym edge} between such pair.
This process enables the KG to link synonyms across different passages, facilitating the integration of both old and new knowledge during learning.
3) Finally, this phrase-based KG is combined with the original passages, allowing the final open KG to incorporate both conceptual and contextual information (\S\ref{subsec:index_integration}).

\noindent \textbf{Online Retrieval.}
1) The query is linked to relevant triples and passages using the encoder, identifying nodes that could be used as seed nodes for graph search (\S \ref{subsec:query_to_triple}).
2) During triple linkage, the recognition memory functions as a filter, ensuring only relevant triples are retained from the retrieved set as the final seed nodes (\S \ref{subsec:recognition_memory}).
3) These final seed nodes are then used to assign reset probabilities within the PPR algorithm, enabling its context-aware retrieval and refining the linking results to retrieve the most relevant passages.
4) Finally, the retrieved passages serve as contextual inputs for the final QA task.
Next, we describe each of the improvements in \ours in more detail.
% by \ours.

\subsection{Dense-Sparse Integration}
\label{subsec:index_integration}

The nodes in the \hipporag KG primarily consist of phrases describing concepts, which we refer to as \textbf{phrase nodes} in this paper. This graph structure introduces limitations related to the \textit{concept-context tradeoff}.
Concepts are concise and easily generalizable but often entail information loss. 
In contrast, context provide specific circumstances that shape the interpretation and application of these concepts, enriching semantics but increasing complexity. 
However, in human memory, concepts and contexts are intricately interconnected.
The dense and sparse coding theory offers insights into how the brain represents and processes information at different granularities \cite{Beyeler2019}. 
Dense coding encodes information through the simultaneous activation of many neurons, resulting in a distributed and \textit{redundant} representation. 
Conversely, sparse coding relies on minimal neural activation, engaging only a small subset of neurons to enhance \textit{efficiency} and storage \textit{compactness}.

Inspired by the dense-sparse integration observed in the human brain, we treat the phrase node as a form of sparse coding for the extracted concepts, while incorporating dense coding into our KG to represent the context from which these concepts originate.
First, we adopt an encoding approach similar to how phrases are encoded, using the embedding model.
These two types of coding are then integrated in a specific manner within the KG.
Unlike the document ensemble in \hipporag, which simply aggregates scores from graph search and embedding matching, we enhance the KG by introducing \textbf{passage nodes}, enabling more seamless integration of contextual information.
This approach retains the same offline indexing process as \hipporag while enriching the graph structure with additional nodes and edges related to passages during construction.
% Specifically, during the graph construction phase, we treat each passage in the corpus as a passage node and add them to the graph. 
% Subsequently, we add directed edges from each passage node to all the phrases the passage contains, where this type of edge is labeled ``contains''. 
Specifically, each passage in the corpus is treated as a passage node, with the \textbf{context edge} labeled ``\textit{contains}'' connecting the passage to all phrases derived from this passage. 
% They are set to be unidirectional edges since the PPR process is undirected.

\subsection{Deeper Contextualization}
\label{subsec:query_to_triple}

Building upon the discussion of the concept-context tradeoff, we observe that query parsing in \hipporag, which relies on Named Entity Recognition (NER), is predominantly concept-centric, often overlooking the contextual alignment within the KG. 
This entity-focused approach to extraction and indexing introduces a strong bias toward concepts, leaving many contextual signals underutilized \cite{hipporag}.
To address this limitation, we explore and evaluate different methods for linking queries to the KG, aiming to more effectively align query semantics with the starting nodes of graph searches. 
Specifically, we consider three approaches:
1) \textbf{NER to Node}: This is the original method used in \hipporag, where entities are extracted from the query and subsequently matched with nodes in the KG using text embeddings.
2) \textbf{Query to Node}: Instead of extracting individual entities, we leverage text embeddings to match the entire query directly to nodes in the KG.
3) \textbf{Query to Triple}: To incorporate richer contextual information from the KG, we match the entire query to triples within the graph using text embeddings. Since triples encapsulate fundamental contextual relationships among concepts, this method provides a more comprehensive understanding of the query's intent.
By default, \ours adopts the query-to-triple approach, and we evaluate all three methods later (\S\ref{subsec:ablation}).

\subsection{Recognition Memory}
\label{subsec:recognition_memory}

Recall and recognition are two complementary processes in human memory retrieval \cite{Uner2022}.
Recall involves actively retrieving information without external cues, while recognition relies on identifying information with the help of external stimuli.
Inspired by this, we model the query-to-triple retrieval as a two-step process.
1) \textbf{Query to Triple}: We use the embedding model to retrieve the top-k triples $T$ of the graph as described in \S\ref{subsec:query_to_triple}. 
2) \textbf{Triple Filtering}: We use LLMs to filter retrieved $T$ and generate triples $T' \subseteq T$. 
The detailed prompts are shown in Appendix \ref{sec:prompts}.

\subsection{Online Retrieval}
\label{subsec:online retrieval}

We summarize the online retrieval process in \ours after introducing the above improvements. 
The task involves selecting seed nodes and assigning reset probabilities for retrieval. \ours identifies phrase nodes from filtered triples generated by query-to-triple and recognition memory. If no triples are available, it directly retrieves top-ranked passages using the embedding model. Otherwise, up to $k$ phrase nodes are selected based on their average ranking scores across filtered triples they originate. 
All passage nodes are also taken as seed nodes, as broader activation improves multi-hop reasoning. 
Reset probabilities are assigned based on ranking scores for phrase nodes, while passage nodes receive scores proportional to their embedding similarity, adjusted by a weight factor (\S\ref{subsec:hyperparameter}) to balance the influence between phrase nodes and passage nodes. 
The PPR search is then executed, and passages are ranked by their PageRank scores, with the top-ranked passages used for downstream QA. An example of the pipeline is in Appendix \ref{sec:pipeline example} and the PPR initialization is detailed in Appendix \ref{PPR_init}.

\begin{table*}[tb]
  \centering
  \small
    \caption{Dataset statistics}
    \vskip 0.1in
  \begin{tabular}{lrrrrrrr}
  \toprule
   & NQ & PopQA & MuSiQue & 2Wiki & HotpotQA & LV-Eval & NarrativeQA   \\ \midrule
  Num of queries & $1,000$ & $1,000$ & $1,000$ & $1,000$ & $1,000$ & $124$ & $293$ \\
  Num of passages & $9,633$ & $8,676$ & $11,656$ & $6,119$ & $9,811$ & $22,849$ & $4,111$ \\
  \bottomrule
  \end{tabular}
  \label{table:dataset statistics}
  % \vskip -0.1in
\end{table*}

\section{Experimental Setup}
\label{section:experimental setup}

\subsection{Baselines}

We select three different types of baselines for comparison. We include three simple baselines: the classic \textbf{BM25} \cite{bm25} baseline as well as \textbf{Contriever} \cite{contriever} and \textbf{GTR} \cite{gtr}, two popular dense embedding retrievers. 

Our second baseline category includes some of the largest embedding models available (7B) that demonstrate strong performance on the BEIR leaderboard \cite{beir}: \textbf{Alibaba-NLP/GTE-Qwen2-7B-Instruct}~\cite{gte}, \textbf{GritLM/GritLM-7B}~\cite{gritlm}, and \textbf{nvidia/NV-Embed-v2}~\cite{nvembedv2}.

In our final baseline category, we include four structure-augmented RAG methods. \textbf{RAPTOR} \cite{raptor} organizes the retrieval corpus into a hierarchical structure based on semantic similarity. \textbf{GraphRAG} \cite{graphrag} and \textbf{LightRAG} \cite{lightrag} leverage a KG structure like ours to generate high-level summaries of the concepts present in the corpus. Finally, \textbf{\hipporag} \cite{hipporag} uses a KG as well but integrates knowledge using PPR rather than summarization.

\subsection{Datasets}

To evaluate how well RAG systems retain factual memory while enhancing associativity and sense-making, we select datasets that correspond to three critical challenge types.
\begin{enumerate}
\item \textbf{Simple QA} primarily evaluates the ability to recall and retrieve factual knowledge accurately.
\item \textbf{Multi-hop QA} measures associativity by requiring the model to connect multiple pieces of information to derive an answer.
\item \textbf{Discourse understanding} evaluates sense-making by testing the capability to interpret and reason over lengthy, complex narratives.
\end{enumerate}

We will now list the datasets chosen for each category and describe them in detail. The statistics for our sampled datasets are summarized in Table \ref{table:dataset statistics}.

\noindent \textbf{Simple QA.}
This common type of QA task primarily involves questions centered around individual entities, making it particularly well-suited for embedding models to retrieve relevant contextual information intuitively.
We randomly collect $1,000$ queries from the \textbf{NaturalQuestions} (NQ) dataset (collected by \citet{rear}), which contains real user questions with a wide range of topics. 
Additionally, we select $1,000$ queries from \textbf{PopQA} \cite{popqa}, with the corpus derived from the December 2021 Wikipedia dump.\footnote{\url{https://github.com/facebookresearch/atlas?tab=readme-ov-file\#corpora}}
Both datasets offer straightforward QA pairs, enabling evaluation of single-hop QA capabilities in RAG systems. 
Notably, PopQA from Wikipedia is especially entity-centric, with entities being less frequent than NaturalQuestions, making it an excellent resource for evaluating entity recognition and retrieval in simple QA tasks.

\noindent \textbf{Multi-hop QA.}
We randomly collect $1,000$ queries from \textbf{MuSiQue}, \textbf{2WikiMultihopQA}, and \textbf{HotpotQA} following HippoRAG \cite{hipporag}, all requiring multi-passage reasoning. 
Additionally, we include all $124$ queries from \textbf{LV-Eval} (hotpotwikiqa-mixup 256k) \cite{lveval}, a challenging dataset designed to minimize knowledge leakage and reduce overfitting through keyword and phrase replacements. 
Thus, unlike Wikipedia-based datasets, LV-Eval better evaluates the model's ability to synthesize knowledge from different sources effectively. 
For corpus collection, we segment long-form contexts of LV-Eval into shorter passages while maintaining the same RAG setup as other multi-hop datasets.

\noindent \textbf{Discourse Understanding.}
This category consists of only \textbf{NarrativeQA}, a QA dataset that contains questions requiring a cohesive understanding of a full-length novel. This dataset's focus on large-scale discourse understanding allows us to leverage it in our evaluation of sense-making in our chosen baselines and our own method. 
We randomly select $10$ lengthy documents and their corresponding $293$ queries from NarrativeQA and collect a retrieval corpus just as in the above LV-Eval dataset.

\begin{table*}[!t]
  \centering
  \small
\caption{\textbf{QA performance} (F1 scores) on RAG benchmarks using Llama-3.3-70B-Instruct as the QA reader. No retrieval means evaluating the parametric knowledge of the readers. All structure-augmented RAG baselines and \ours use Llama-3.3-70B-Instruct as the LLM to generate their structure and NV-Embed-v2 as their retriever. This table, along with the following ones, highlight the \textbf{best} and \underline{second-best} results. A bootstrapped statistical test was used to assess significance; $^{\dagger}$ indicates that \ours significantly outperforms the best NV-Embed-v2 baseline ($p < 0.05$).}
\vskip 0.1in
\resizebox{1.0\textwidth}{!}{
\begin{tabular}{lcccccccc}
\toprule
& \multicolumn{2}{c}{\multirow{2}{*}{Simple QA}} 
& \multicolumn{4}{c}{\multirow{2}{*}{Multi-Hop QA}} 
& \multicolumn{1}{c}{\multirow{2}{*}{\shortstack{Discourse\\Understanding}}} & \\
&&&&&&&&\\
\cmidrule(lr){2-3} \cmidrule(lr){4-7} \cmidrule(lr){8-8}
Retrieval & NQ & PopQA & MuSiQue & 2Wiki & HotpotQA & LV-Eval & NarrativeQA & Avg \\
\midrule
\rowcolor{lightgrey}
\multicolumn{9}{c}{\textit{\textbf{Simple Baselines}}} \\
None & $54.9$ & $32.5$ & $26.1$ & $42.8$ & $47.3$ & \ \ $6.0$ & $12.9$ & $38.4$ \\ 
Contriever~\cite{contriever} & $58.9$ & $53.1$ & $31.3$ & $41.9$ & $62.3$ & \ \ $8.1$ & $19.7$ & $46.9$ \\
BM25~\cite{bm25} & $59.0$ & $49.9$ & $28.8$ & $51.2$ & $63.4$ & \ \ $5.9$ & $18.3$ & $47.7$ \\ 
GTR (T5-base)~\cite{gtr} & $59.9$ & $\underline{56.2}$ & $34.6$ & $52.8$ & $62.8$ & \ \ $7.1$ & $19.9$ & $50.4$\\ 
\rowcolor{lightgrey}
\multicolumn{9}{c}{\textit{\textbf{Large Embedding Models}}} \\
GTE-Qwen2-7B-Instruct~\cite{gte} & $\underline{62.0} $ & $\mathbf{56.3}$ & $40.9$ & $60.0$ & $71.0$ & \ \ $7.1$ & $21.3$ & $54.9$ \\
GritLM-7B~\cite{gritlm} & $61.3$ & $55.8$ & $44.8$ & $60.6$ & $73.3$ & \ \ $9.8$ & $23.9$ & $56.1$ \\ 
NV-Embed-v2 (7B)~\cite{nvembedv2} & $61.9$ & $55.7$ & $\underline{45.7}$ & $61.5$ & $\underline{75.3}$ & \ \ $9.8$ & $\underline{25.7}$ & $\underline{57.0}$ \\
\rowcolor{lightgrey}
\multicolumn{9}{c}{\textit{\textbf{Structure-Augmented RAG}}} \\
RAPTOR~\cite{raptor} & $50.7$ & $\underline{56.2}$ & $28.9$ &  $52.1$  & $69.5$  &  \ \ $5.0$ & $21.4$ & $48.8$ \\
GraphRAG~\cite{graphrag} & $46.9$ &  $48.1$ & $38.5$  &  $58.6$  & $68.6$ & $\underline{11.2}$ & $23.0$ & $49.6$ \\
LightRAG~\cite{lightrag} & $16.6$ & \ \ $2.4$  & \ \ $1.6$ &  $11.6$  & \ \ $2.4$ &  \ \ $1.0$ & \ \ $3.7$ & \ \ $6.6$ \\
HippoRAG~\cite{hipporag} &  $55.3$ &  $55.9$ &  $35.1$ & $\mathbf{71.8}$ &	$63.5$  & \ \ $8.4$ & $16.3$ & $53.1$ \\
\midrule
\ours  & $\mathbf{63.3}$$^{\dagger}$ &  $\underline{56.2}$ &  $\mathbf{48.6}$$^{\dagger}$ & $\underline{71.0}$$^{\dagger}$ & $\mathbf{75.5}$ & $\mathbf{12.9}$$^{\dagger}$ & $\mathbf{25.9}$ & $\mathbf{59.8}$ \\
\bottomrule
\end{tabular}}
\label{table:QA_results}
\vskip -0.1in
\end{table*}

\begin{table*}[!h]
  \centering
  \small
    \caption{\textbf{Retrieval performance} (passage recall@5) on RAG benchmarks. * denotes the report from the original paper. The compared structure-augmented RAG methods are reproduced with the same LLM and retriever as ours for a fair comparison. GraphRAG and LightRAG are not presented because they do not directly produce passage retrieval results. }
\vskip 0.1in
  \begin{tabular}{lcccccc}
  \toprule
    & \multicolumn{2}{c}{Simple QA} & \multicolumn{3}{c}{Multi-Hop QA} \\
  \cmidrule(lr){2-3} \cmidrule(lr){4-6}
  Retrieval & NQ & PopQA & MuSiQue & 2Wiki & HotpotQA & Avg \\
  \midrule
  \rowcolor{lightgrey}
  \multicolumn{7}{c}{\textit{\textbf{Simple Baselines}}} \\
  BM25~\cite{bm25} & $56.1$ & $35.7$ & $43.5$ & $65.3$ & $74.8$ & $55.1$  \\ 
  % \midrule
  Contriever~\cite{contriever} & $54.6$ & $43.2$ & $46.6$ & $57.5$ &  $75.3$ & $55.4$ \\
  GTR (T5-base)~\cite{gtr} & $63.4$ & $49.4$ & $49.1$ & $67.9$ & $73.9$ & $60.7$ \\
  \rowcolor{lightgrey}
  \multicolumn{7}{c}{\textit{\textbf{Large Embedding Models}}} \\
  GTE-Qwen2-7B-Instruct~\cite{gte} & $74.3$ & $50.6$ & $63.6$ & $74.8$ & $89.1$ & $70.5$  \\
  GritLM-7B~\cite{gritlm} & $\underline{76.6}$ & $50.1$ &  $65.9$ &  $76.0$ & $92.4$ & $72.2$ \\
  NV-Embed-v2 (7B)~\cite{nvembedv2} & $75.4$ & $51.0$ & $\underline{69.7}$ & $76.5$ & $\underline{94.5}$ & $\underline{73.4}$  \\ 
  \rowcolor{lightgrey}
  \multicolumn{7}{c}{\textit{\textbf{Structure-Augmented RAG}}} \\
  % RAPTOR~\cite{raptor} &$40.5$ / $69.4$ & $37.2$ / $48.1$ & $49.1$ / $61.0$ & $61.5$ / $70.6$ & $78.6$ / $90.2$  \\
  RAPTOR~\cite{raptor} & $68.3$  &  $48.7$  &  $57.8$ &  $66.2$ & $86.9$ & $65.6$  \\
  % GraphRAG & & & & & & & \\
  % LightRAG & & & & & & & \\
  HippoRAG*~\cite{hipporag} & $-$ & $-$ &  $51.9$ & $\underline{89.1}$ & $77.7$ & $-$ \\
  % HippoRAG~(GPT-4o-mini) & 45.1 & 52.2 & $52.4$ & $87.0$ & $78.5$ & 63.0 \\ 
  HippoRAG (reproduced) &  $44.4$ & $\mathbf{53.8}$ & $53.2$ &  $\mathbf{90.4}$ &  $77.3$ & $63.8$\\
   % KAG*~\cite{} & - & - & 48.5 / 65.7 & 65.4 / 91.9 & - \\
   \midrule
  % \ours (GPT-4o-mini) &  76.4 &  52.2 & 74.2 &  90.2 &  95.7 & 77.7   \\
  \ours &  $\mathbf{78.0}$ &  $\underline{51.7}$ & $\mathbf{74.7}$ & $\mathbf{90.4}$ & $\mathbf{96.3}$ & $\mathbf{78.2}$ \\
  % \ours & & & $\mathbf{55.4}$ / $\mathbf{73.3}$ & $\mathbf{74.1}$ / $\mathbf{87.4}$ & $\mathbf{84.9}$ / $\mathbf{95.9}$ % 4o filter\\
  \bottomrule
  \end{tabular}
  % \vspace{-10pt}
  \label{table:recall_results}
\vskip -0.1in
\end{table*}

\subsection{Metrics}

Following HippoRAG \cite{hipporag}, we use passage recall@5 to evaluate the retrieval task.
For the QA task, we follow evaluation metrics from MuSiQue \cite{musique} to calculate token-based F1 scores.

\subsection{Implementation Details}
\label{subsec: implementation details}

For \ours, we use the open-source Llama-3.3-70B-Instruct \cite{llama3modelcard} as both the extraction (NER and OpenIE) and triple filtering model, and we use nvidia/NV-Embed-v2 as the retriever.
We also reproduce the compared structure-augmented RAG methods using the same extractor and retriever for a fair comparison.
For the triple filter, we use DSPy \cite{dspy} MIPROv2 optimizer and Llama-3.3-70B-Instruct to tune the prompt, including the instructions and demonstrations. 
The resulting prompt is shown in Appendix \ref{sec:prompts}.
We use top-5 triples ranked by retriever for filtering.
Our QA module uses the top-5 retrieved passages as context for an LLM (GPT-4o-mini or Llama-3.3-70B-Instruct) to generate the final answer. 
For hyperparameters, we follow the default settings from \hipporag. 
More implementation and hyperparameter details can be found in Appendix \ref{sec:appendix-hyper}.

\section{Results}

We now present our main QA and retrieval experimental results, where the QA process uses retrieved results as its context. 
More detailed experimental results are presented in Appendix \ref{sec:detailed results}. The statistics for all constructed KGs are shown in Appendix \ref{sec:prompts}.

\noindent \textbf{QA Performance.} Table \ref{table:QA_results} presents the QA performance of various retrievers across multiple RAG benchmarks using Llama-3.3-70B-Instruct as the QA reader. \ours achieves the highest average F1 score, demonstrating robustness across different settings. 
Large embedding models outperform smaller ones, with NV-Embed-v2 (7B) scoring $6.6\%$ higher on average than GTR (T5-base). 
These models also surpass structure-augmented RAG methods with lower computational costs but excel mainly in simple QA while struggling in complex cases. Notably, \ours outperforms NV-Embed-v2 by $9.5\%$ F1 on 2Wiki and by $3.1\%$ on the challenging LV-Eval dataset. Compared to \hipporag, \ours shows even greater improvements, validating its neuropsychology-inspired approach. These results highlight \ours as a state-of-the-art RAG system that enhances both retrieval and QA performance while being effectively powered by an open-source model.
Table \ref{table:appendix qa performance} in Appendix \ref{sec:detailed results} presents additional QA results (EM and F1) using Llama or GPT-4o-mini as the QA reader, along with an extractor or triple filter. GPT-4o-mini follows Llama’s trend, with NV-Embed-v2 outperforming structure-augmented methods in most cases, except for HippoRAG in multi-hop QA. \ours consistently outperforms all other methods across nearly all settings. An analysis of the computational resources (tokens, time and memory) required for each method can be found in Appendix \ref{sec:cost and efficiency}.

\noindent \textbf{Retrieval Performance.} We report retrieval results for datasets with supporting passage annotations and models that explicitly retrieve passages in Table \ref{table:recall_results}. Large embedding models (7B) significantly outperform classic smaller LM-based models like Contriever and GTR, achieving at least a $9.8\%$ higher F1 score. While our reproduction of HippoRAG using Llama-3.3-70B-Instruct and NV-Embed-v2 shows slight improvements over the original paper, the gains are minimal, with only a $1.3\%$ increase in F1. Although \hipporag excels in entity-centric retrieval, achieving the highest recall@5 on PopQA, it generally lags behind recent dense retrievers and \ours. Notably, \ours achieves the highest recall scores across most datasets, with substantial improvements of $5.0\%$ and $13.9\%$ in Recall@5 on MuSiQue and 2Wiki, respectively, compared to the strongest dense retriever, NV-Embed-v2.

\begin{table}[t]
  \centering
  \small
\caption{\textbf{Ablations.} We report passage recall@5 on multi-hop QA benchmarks using several alternatives to our final design in graph linking, graph construction and triple filtering.}
\vskip 0.1in
  \resizebox{\columnwidth}{!}{%
  \begin{tabular}{lcccccccc}
  \toprule
   & MuSiQue & 2Wiki & HotpotQA & Avg \\
  \midrule
  \ours &  $\mathbf{74.7}$ &  $90.4$ &  $\mathbf{96.3}$ &  $\mathbf{87.1}$ \\  
  \midrule 
  \ \ w/ NER to node &  $53.8$ &  $\mathbf{91.2}$ & $78.8$ & $74.6$ \\
  \ \ w/ Query to node &  $44.9$ &  $65.5$ & $68.3$ &  $59.6$ \\  
  \midrule
  \ \ w/o Passage Node & $63.7$ & $90.3$ & $88.9$ & $81.0$ \\ 
  \midrule
  \ \ w/o Filter & $\underline{73.0}$ & $\underline{90.7}$ & $\underline{95.4}$ & $
  \underline{86.4}$ \\
  \bottomrule
  \end{tabular}%
  }
  % \vspace{-10pt}
  \label{table:ablations}
  \vskip -0.1in
\end{table}

\begin{table}[t]
\centering
\small
\caption{\textbf{Reset probability factor.} Passage recall@5 with different weight factors for passage nodes on our MuSiQue dev set and NaturalQuestions (NQ) dev set, where each set has $1,000$ queries.}
\vskip 0.1in
\begin{tabular}{lcccccc}
\toprule
Weight & $0.01$ & $0.05$ & $0.1$ & $0.3$ & $0.5$ \\ \midrule
MuSiQue & $79.9$ & $\mathbf{80.5}$ & $79.8$ & $78.4$ & $77.9$ \\ 
NQ & $75.6$ & $\mathbf{76.9}$ & $76.9$ & $76.7$ & $76.4$ \\
\bottomrule
\end{tabular}%
% }
\label{table:weight factor}
\vskip -0.1in
\end{table}

\section{Discussions}

\subsection{Ablation Study}
\label{subsec:ablation}
We design ablation experiments for the proposed linking method, graph construction method, and triple filtering method, with the results reported in Table \ref{table:ablations}.
Each introduced mechanism boosts \ours.
First, the linking method with deeper contextualization leads to significant performance improvements. 
Notably, we do not apply a filtering process to the NER-to-node or query-to-node methods; however, the query-to-triple approach, regardless of whether filtering is applied, consistently outperforms the other two linking strategies. On average, query-to-triple improves Recall@5 by $12.5\%$ compared to NER-to-node. Moreover, query-to-node does not provide an advantage over NER-to-node, as queries and KG nodes operate at different levels of granularity, whereas both NER results and KG nodes correspond to phrase-level representations.

\begin{figure}[t]
    \centering
    \includegraphics[width=\columnwidth]{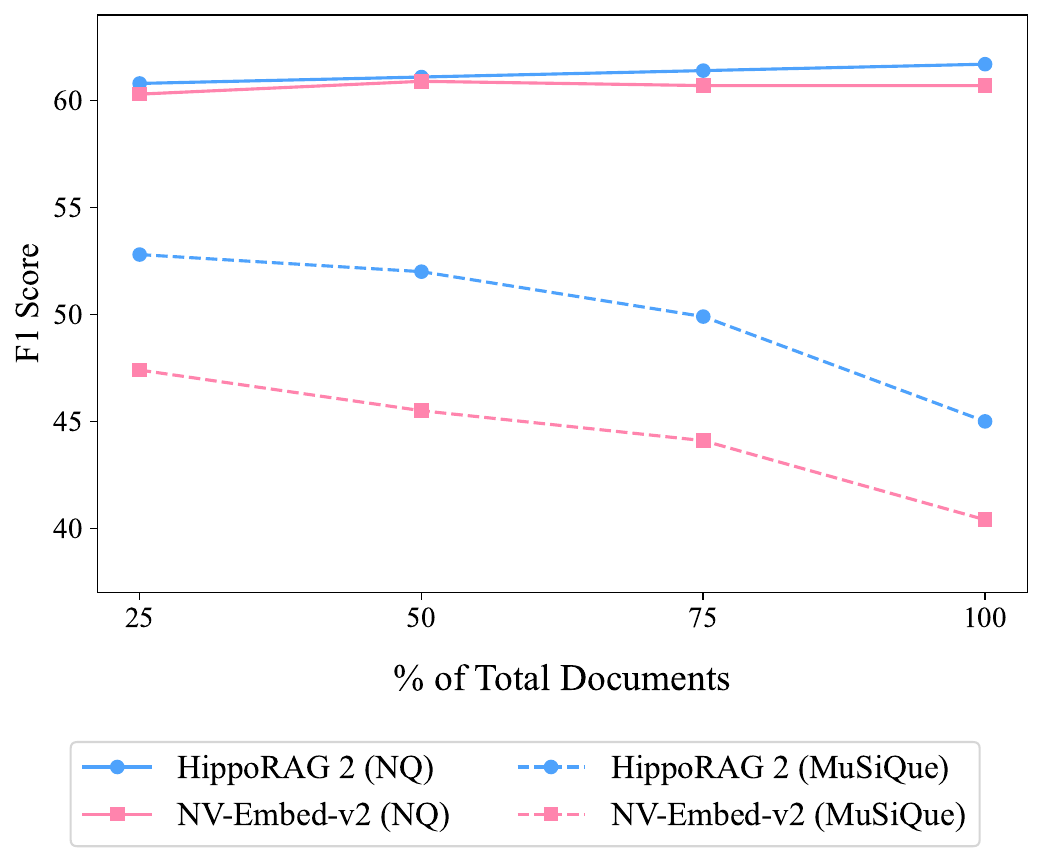}
      \vskip -0.1in
    \caption{\textbf{Continual learning experiment:} We partition the NQ and MuSiQue datasets into 4 segments and report the F1 score on a randomly chosen segment as the other 3 segments are introduced into the retrieval corpus to simulate a continuously evolving corpus.}
    \label{continual}
  \vskip -0.2in
\end{figure}

\subsection{Controlling Reset Probabilities}
\label{subsec:hyperparameter}

When setting the reset probability before starting PPR, we find that it is necessary to balance the reset probabilities between two types of nodes: phrase nodes and passage nodes. 
Specifically, the reset probability of all passage nodes is multiplied by a weight factor to balance the importance of two types of nodes during PPR.
Here, we present the results obtained on the validation set in Table \ref{table:weight factor}, which shows that this factor is crucial for the PPR results.
Considering the model performance across different scenarios, we set the factor to be $0.05$ by default.

\subsection{Robustness to Corpus Expansion}

As RAG systems become more widely adopted in the real-world, they must increasingly adapt to continual learning scenarios in which the retrieval corpora grow continuously. To understand how \ours's capacity to handle this setting compared to standard RAG, we design an experiment in which we partition NQ and MuSiQue into four equal segments, each containing the gold documents and distractors for approximately 250 questions. We then select one segment for evaluation and incrementally add the remaining segments, measuring how performance evolves as new knowledge is added, allowing us to simulate a continual learning setting. We show the F1 score for \ours and NV-Embed-v2, our strongest baseline, in Figure \ref{continual}.

\begin{table*}[htb]
\centering
\small
 \caption{We show exemplary retrieval results (the title of passages) from \ours and NV-Embed-v2 on different types of questions. Bolded items denote the titles of supporting passages.}
 \vskip 0.1in
 \resizebox{\textwidth}{!}{%
\begin{tabular}{p{1.7cm} p{2.2cm} p{3.7cm} p{4.0cm} p{3.7cm}} 
\toprule
& \textbf{Question} & \textbf{NV-Embed-v2} Results & \textbf{\ours} Filtered Triples & \textbf{\ours} Results   \\
\midrule
\textbf{Simple QA} & In what city was I.P. Paul born? & \textbf{1. I. P. Paul}  \newline 2. Yinka Ayefele - Early life \newline  3. Paul Parker (singer) & (\textbf{I. P. Paul}, from, \textbf{Thrissur}) \newline (\textbf{I. P. Paul}, was mayor of, Thrissur municipal corporation) & \textbf{1. I. P. Paul} \newline \textbf{2. Thrissur} \newline 3. Yinka Ayefele \\ 
\midrule
\textbf{Multi-Hop QA} & What county is Erik Hort's birthplace a part of? & \textbf{1. Erik Hort} \newline 2. Horton Park (Saint Paul, Minnesota) \newline 3. Hertfordshire  & (\textbf{Erik Hort}, born in, \textbf{Montebello}) \newline (\textbf{Erik Hort}, born in, New York) &  \textbf{1. Erik Hort} \newline 2. Horton Park (Saint Paul, Minnesota) \newline \textbf{3. Monstebello, New York} \\
\bottomrule
\end{tabular}
}
   \label{table:qualitative analysis}
\vskip -0.1in
\end{table*}

\begin{table}[t]
\centering
\caption{\textbf{Robust to different dense retrievers.} Passage recall@5 on MuSiQue subset.}
\vskip 0.1in
\small
\resizebox{\columnwidth}{!}{%
\begin{tabular}{lcc}
\toprule
Retriever & Dense Retrieval & \ours \\ \midrule
GTE-Qwen2-7B-Instruct   &  $63.6$ & $\mathbf{68.8}$ \\
GritLM-7B               &  $66.0$ & $\mathbf{71.6}$ \\
NV-Embed-v2 (7B)        &  $69.7$ & $\mathbf{74.7}$ \\
\bottomrule
\end{tabular}
}
\label{table:dense retriever flexibility}
\vskip -0.1in
\end{table}

As we can see in Figure \ref{continual}, \ours's improvements over NV-Embed-v2 remain remarkably consistent in both simple (NQ) and associative (MuSiQue) continual learning settings. We also note that, while both methods retain strong performance on simple QA (solid lines) as more knowledge is introduced, their performance in the more complex associative task (dotted lines) degrades at a similar rate as more information is introduced. This divergence underscores the importance of incorporating varied task complexities into future continual learning benchmarks.

\subsection{Dense Retriever Flexibility}

As demonstrated in Table \ref{table:dense retriever flexibility}, \ours consistently surpasses direct dense retrieval across various retrievers.  
Notably, these performance gains remain robust regardless of the specific dense retriever used.

\subsection{Qualitative Analysis}

We show examples from PopQA and MuSiQue in Table \ref{table:qualitative analysis}.
For the first example, ``\textit{In what city was I. P. Paul born?}'', NV-Embed-v2 ranks the entity mentioned in the query ``\textit{I. P. Paul}'' as the top 1, where the passage is enough to answer this question.
But \ours does even better. It directly finds the answer ``\textit{Thrissur}'' when linking the triples, and during the subsequent graph search, it places the passage corresponding to that entity in the second position, which is a perfect retrieval result.
For the second multi-hop question, ``\textit{What county is Erik Hort’s birthplace a part of?}'' NV-Embed-v2 also easily identifies the person mentioned, ``\textit{Erik Hort}.'' 
However, since this question requires two-step reasoning, it is not sufficient to fully answer the question. 
In contrast, \ours retrieves a passage titled ``\textit{Montebello}'' during the query-to-triple step, which contains geographic information that implies the answer to the question. In the subsequent graph search, this passage is also ranked at the top.
Apart from this, the error analysis of \ours is detailed in Appendix \ref{sec:error analysis}.

\section{Conclusion}

We introduced \ours, a novel framework designed to address the limitations of existing RAG systems in approximating the dynamic and interconnected nature of human long-term memory. It combining the strengths of the Personalized PageRank algorithm, deeper passage integration, and effective online use of LLMs. 
\ours opens new avenues for research in continual learning and long-term memory for LLMs by achieving comprehensive improvements over standard RAG methods across factual, sense-making, and associative memory tasks, showing capabilities that previous methods have either overlooked or been incapable of achieving in a thorough evaluation.
% , moving closer to mimicking the adaptability and efficiency of human learning. 
Future work could consider leveraging graph-based retrieval methods to further enhance the episodic memory capabilities of LLMs in long conversations.

\section*{Impact Statement}

This paper presents work on Retrieval-Augmented Generation (RAG) to advance the field of long-term memory for large language models. 
While our work may have various societal implications, we do not identify any concerns that warrant specific emphasis beyond those generally associated with large language models and information retrieval systems.

\section*{Acknowledgments}
 
We would also like to extend our appreciation to colleagues from the OSU NLP group for their constructive comments. 
This work is supported in part by ARL W911NF2220144, NSF 2112606, and a gift from Cisco. 
We also thank the Ohio Supercomputer Center for providing computational resources.
The views and conclusions contained herein are those of the
authors and should not be interpreted as representing the official policies, either expressed or implied,
of the U.S.\ government. The U.S.\ government is
authorized to reproduce and distribute reprints for
government purposes notwithstanding any copyright notice herein.


\bibliography{anthology, custom}

\begin{thebibliography}{43}
\providecommand{\natexlab}[1]{#1}
\providecommand{\url}[1]{\texttt{#1}}
\expandafter\ifx\csname urlstyle\endcsname\relax
  \providecommand{\doi}[1]{doi: #1}\else
  \providecommand{\doi}{doi: \begingroup \urlstyle{rm}\Url}\fi

\bibitem[AI@Meta(2024)]{llama3modelcard}
AI@Meta.
\newblock Llama 3 model card.
\newblock 2024.
\newblock URL \url{https://github.com/meta-llama/llama3/blob/main/MODEL_CARD.md}.

\bibitem[Beyeler et~al.(2019)Beyeler, Rounds, Carlson, Dutt, and Krichmar]{Beyeler2019}
Beyeler, M., Rounds, E.~L., Carlson, K.~D., Dutt, N., and Krichmar, J.~L.
\newblock Neural correlates of sparse coding and dimensionality reduction.
\newblock \emph{PLoS Comput Biol}, 15\penalty0 (6):\penalty0 e1006908, 2019.
\newblock \doi{10.1371/journal.pcbi.1006908}.

\bibitem[Chen et~al.(2023)Chen, Pasunuru, Weston, and Celikyilmaz]{chen2023walkingmemorymazecontext}
Chen, H., Pasunuru, R., Weston, J., and Celikyilmaz, A.
\newblock Walking down the memory maze: Beyond context limit through interactive reading, 2023.
\newblock URL \url{https://arxiv.org/abs/2310.05029}.

\bibitem[Cohen et~al.(2024)Cohen, Biran, Yoran, Globerson, and Geva]{cohen-etal-2024-evaluating}
Cohen, R., Biran, E., Yoran, O., Globerson, A., and Geva, M.
\newblock Evaluating the ripple effects of knowledge editing in language models.
\newblock \emph{Transactions of the Association for Computational Linguistics}, 12:\penalty0 283--298, 2024.
\newblock \doi{10.1162/tacl_a_00644}.
\newblock URL \url{https://aclanthology.org/2024.tacl-1.16/}.

\bibitem[Edge et~al.(2024)Edge, Trinh, Cheng, Bradley, Chao, Mody, Truitt, and Larson]{graphrag}
Edge, D., Trinh, H., Cheng, N., Bradley, J., Chao, A., Mody, A., Truitt, S., and Larson, J.
\newblock From local to global: A graph rag approach to query-focused summarization, 2024.
\newblock URL \url{https://arxiv.org/abs/2404.16130}.

\bibitem[Gu et~al.(2024)Gu, Xu, Ma, Lu, Ling, Chang, and Peng]{gu-etal-2024-model}
Gu, J.-C., Xu, H.-X., Ma, J.-Y., Lu, P., Ling, Z.-H., Chang, K.-W., and Peng, N.
\newblock Model editing harms general abilities of large language models: Regularization to the rescue.
\newblock In Al-Onaizan, Y., Bansal, M., and Chen, Y.-N. (eds.), \emph{Proceedings of the 2024 Conference on Empirical Methods in Natural Language Processing}, pp.\  16801--16819, Miami, Florida, USA, November 2024. Association for Computational Linguistics.
\newblock \doi{10.18653/v1/2024.emnlp-main.934}.
\newblock URL \url{https://aclanthology.org/2024.emnlp-main.934/}.

\bibitem[Guo et~al.(2024)Guo, Xia, Yu, Ao, and Huang]{lightrag}
Guo, Z., Xia, L., Yu, Y., Ao, T., and Huang, C.
\newblock {LightRAG}: Simple and fast retrieval-augmented generation, 2024.
\newblock URL \url{https://arxiv.org/abs/2410.05779}.

\bibitem[Gutiérrez et~al.(2024)Gutiérrez, Shu, Gu, Yasunaga, and Su]{hipporag}
Gutiérrez, B.~J., Shu, Y., Gu, Y., Yasunaga, M., and Su, Y.
\newblock Hipporag: Neurobiologically inspired long-term memory for large language models.
\newblock In \emph{The Thirty-eighth Annual Conference on Neural Information Processing Systems}, 2024.
\newblock URL \url{https://openreview.net/forum?id=hkujvAPVsg}.

\bibitem[Haveliwala(2002)]{david02topic}
Haveliwala, T.~H.
\newblock Topic-sensitive pagerank.
\newblock In Lassner, D., Roure, D.~D., and Iyengar, A. (eds.), \emph{Proceedings of the Eleventh International World Wide Web Conference, {WWW} 2002, May 7-11, 2002, Honolulu, Hawaii, {USA}}, pp.\  517--526. {ACM}, 2002.
\newblock \doi{10.1145/511446.511513}.
\newblock URL \url{https://dl.acm.org/doi/10.1145/511446.511513}.

\bibitem[Hoelscher-Obermaier et~al.(2023)Hoelscher-Obermaier, Persson, Kran, Konstas, and Barez]{hoelscher-obermaier-etal-2023-detecting}
Hoelscher-Obermaier, J., Persson, J., Kran, E., Konstas, I., and Barez, F.
\newblock Detecting edit failures in large language models: An improved specificity benchmark.
\newblock In Rogers, A., Boyd-Graber, J., and Okazaki, N. (eds.), \emph{Findings of the Association for Computational Linguistics: ACL 2023}, pp.\  11548--11559, Toronto, Canada, July 2023. Association for Computational Linguistics.
\newblock \doi{10.18653/v1/2023.findings-acl.733}.
\newblock URL \url{https://aclanthology.org/2023.findings-acl.733/}.

\bibitem[Huang et~al.(2024)Huang, Cui, Wang, Yang, Liao, Song, Yao, and Su]{huang24mitigating}
Huang, J., Cui, L., Wang, A., Yang, C., Liao, X., Song, L., Yao, J., and Su, J.
\newblock Mitigating catastrophic forgetting in large language models with self-synthesized rehearsal.
\newblock In Ku, L.-W., Martins, A., and Srikumar, V. (eds.), \emph{Proceedings of the 62nd Annual Meeting of the Association for Computational Linguistics (Volume 1: Long Papers)}, pp.\  1416--1428, Bangkok, Thailand, August 2024. Association for Computational Linguistics.
\newblock \doi{10.18653/v1/2024.acl-long.77}.
\newblock URL \url{https://aclanthology.org/2024.acl-long.77/}.

\bibitem[Izacard et~al.(2022)Izacard, Caron, Hosseini, Riedel, Bojanowski, Joulin, and Grave]{contriever}
Izacard, G., Caron, M., Hosseini, L., Riedel, S., Bojanowski, P., Joulin, A., and Grave, E.
\newblock Unsupervised dense information retrieval with contrastive learning.
\newblock \emph{Trans. Mach. Learn. Res.}, 2022, 2022.
\newblock URL \url{https://openreview.net/forum?id=jKN1pXi7b0}.

\bibitem[Jin et~al.(2022)Jin, Zhang, Zhu, Xiao, Li, Wei, Arnold, and Ren]{lifelong}
Jin, X., Zhang, D., Zhu, H., Xiao, W., Li, S.-W., Wei, X., Arnold, A., and Ren, X.
\newblock Lifelong pretraining: Continually adapting language models to emerging corpora.
\newblock In Carpuat, M., de~Marneffe, M.-C., and Meza~Ruiz, I.~V. (eds.), \emph{Proceedings of the 2022 Conference of the North American Chapter of the Association for Computational Linguistics: Human Language Technologies}, pp.\  4764--4780, Seattle, United States, July 2022. Association for Computational Linguistics.
\newblock \doi{10.18653/v1/2022.naacl-main.351}.
\newblock URL \url{https://aclanthology.org/2022.naacl-main.351/}.

\bibitem[Khattab et~al.(2024)Khattab, Singhvi, Maheshwari, Zhang, Santhanam, Vardhamanan, Haq, Sharma, Joshi, Moazam, Miller, Zaharia, and Potts]{dspy}
Khattab, O., Singhvi, A., Maheshwari, P., Zhang, Z., Santhanam, K., Vardhamanan, S., Haq, S., Sharma, A., Joshi, T.~T., Moazam, H., Miller, H., Zaharia, M., and Potts, C.
\newblock {DSPy}: Compiling declarative language model calls into self-improving pipelines.
\newblock 2024.

\bibitem[Kim et~al.(2024)Kim, Yoon, Ye, Bae, Ho, Hwang, and Yun]{kim-etal-2024-carpe}
Kim, Y., Yoon, J., Ye, S., Bae, S., Ho, N., Hwang, S.~J., and Yun, S.-Y.
\newblock Carpe diem: On the evaluation of world knowledge in lifelong language models.
\newblock In Duh, K., Gomez, H., and Bethard, S. (eds.), \emph{Proceedings of the 2024 Conference of the North American Chapter of the Association for Computational Linguistics: Human Language Technologies (Volume 1: Long Papers)}, pp.\  5401--5415, Mexico City, Mexico, June 2024. Association for Computational Linguistics.
\newblock \doi{10.18653/v1/2024.naacl-long.302}.
\newblock URL \url{https://aclanthology.org/2024.naacl-long.302/}.

\bibitem[Klein et~al.(2006)Klein, Moon, and Hoffman]{klein2006making}
Klein, G., Moon, B., and Hoffman, R.~R.
\newblock Making sense of sensemaking 1: Alternative perspectives.
\newblock \emph{IEEE intelligent systems}, 21\penalty0 (4):\penalty0 70--73, 2006.

\bibitem[Koli et~al.(2024)Koli, Yuan, and Dasgupta]{koli-etal-2024-sensemaking}
Koli, V., Yuan, J., and Dasgupta, A.
\newblock Sensemaking of socially-mediated crisis information.
\newblock In Blodgett, S.~L., Cercas~Curry, A., Dev, S., Madaio, M., Nenkova, A., Yang, D., and Xiao, Z. (eds.), \emph{Proceedings of the Third Workshop on Bridging Human--Computer Interaction and Natural Language Processing}, pp.\  74--81, Mexico City, Mexico, June 2024. Association for Computational Linguistics.
\newblock \doi{10.18653/v1/2024.hcinlp-1.7}.
\newblock URL \url{https://aclanthology.org/2024.hcinlp-1.7/}.

\bibitem[Kwon et~al.(2023)Kwon, Li, Zhuang, Sheng, Zheng, Yu, Gonzalez, Zhang, and Stoica]{vllm}
Kwon, W., Li, Z., Zhuang, S., Sheng, Y., Zheng, L., Yu, C.~H., Gonzalez, J.~E., Zhang, H., and Stoica, I.
\newblock Efficient memory management for large language model serving with pagedattention.
\newblock In \emph{Proceedings of the ACM SIGOPS 29th Symposium on Operating Systems Principles}, 2023.

\bibitem[Lee et~al.(2025)Lee, Roy, Xu, Raiman, Shoeybi, Catanzaro, and Ping]{nvembedv2}
Lee, C., Roy, R., Xu, M., Raiman, J., Shoeybi, M., Catanzaro, B., and Ping, W.
\newblock {NV}-embed: Improved techniques for training {LLM}s as generalist embedding models.
\newblock In \emph{The Thirteenth International Conference on Learning Representations}, 2025.
\newblock URL \url{https://openreview.net/forum?id=lgsyLSsDRe}.

\bibitem[Li et~al.(2024)Li, Armandpour, Mirzadeh, Mehta, Shankar, Vemulapalli, Tuzel, Farajtabar, Pouransari, and Faghri]{ticlm}
Li, J., Armandpour, M., Mirzadeh, S.~I., Mehta, S., Shankar, V., Vemulapalli, R., Tuzel, O., Farajtabar, M., Pouransari, H., and Faghri, F.
\newblock Tic-{LM}: A multi-year benchmark for continual pretraining of language models.
\newblock In \emph{NeurIPS 2024 Workshop on Scalable Continual Learning for Lifelong Foundation Models}, 2024.
\newblock URL \url{https://openreview.net/forum?id=PpSDVE5rAy}.

\bibitem[Li et~al.(2023)Li, Zhang, Zhang, Long, Xie, and Zhang]{gte}
Li, Z., Zhang, X., Zhang, Y., Long, D., Xie, P., and Zhang, M.
\newblock Towards general text embeddings with multi-stage contrastive learning.
\newblock \emph{arXiv preprint arXiv:2308.03281}, 2023.

\bibitem[Liska et~al.(2022)Liska, Kocisky, Gribovskaya, Terzi, Sezener, Agrawal, De~Masson~D'Autume, Scholtes, Zaheer, Young, Gilsenan-Mcmahon, Austin, Blunsom, and Lazaridou]{streamingqa}
Liska, A., Kocisky, T., Gribovskaya, E., Terzi, T., Sezener, E., Agrawal, D., De~Masson~D'Autume, C., Scholtes, T., Zaheer, M., Young, S., Gilsenan-Mcmahon, E., Austin, S., Blunsom, P., and Lazaridou, A.
\newblock {S}treaming{QA}: A benchmark for adaptation to new knowledge over time in question answering models.
\newblock In Chaudhuri, K., Jegelka, S., Song, L., Szepesvari, C., Niu, G., and Sabato, S. (eds.), \emph{Proceedings of the 39th International Conference on Machine Learning}, volume 162 of \emph{Proceedings of Machine Learning Research}, pp.\  13604--13622. PMLR, 17--23 Jul 2022.
\newblock URL \url{https://proceedings.mlr.press/v162/liska22a.html}.

\bibitem[Lù(2024)]{bm25s}
Lù, X.~H.
\newblock {BM25S}: Orders of magnitude faster lexical search via eager sparse scoring, 2024.
\newblock URL \url{https://arxiv.org/abs/2407.03618}.

\bibitem[Mallen et~al.(2023)Mallen, Asai, Zhong, Das, Khashabi, and Hajishirzi]{popqa}
Mallen, A., Asai, A., Zhong, V., Das, R., Khashabi, D., and Hajishirzi, H.
\newblock When not to trust language models: Investigating effectiveness of parametric and non-parametric memories.
\newblock In Rogers, A., Boyd-Graber, J., and Okazaki, N. (eds.), \emph{Proceedings of the 61st Annual Meeting of the Association for Computational Linguistics (Volume 1: Long Papers)}, pp.\  9802--9822, Toronto, Canada, July 2023. Association for Computational Linguistics.
\newblock \doi{10.18653/v1/2023.acl-long.546}.
\newblock URL \url{https://aclanthology.org/2023.acl-long.546/}.

\bibitem[Muennighoff et~al.(2025)Muennighoff, SU, Wang, Yang, Wei, Yu, Singh, and Kiela]{gritlm}
Muennighoff, N., SU, H., Wang, L., Yang, N., Wei, F., Yu, T., Singh, A., and Kiela, D.
\newblock Generative representational instruction tuning.
\newblock In \emph{The Thirteenth International Conference on Learning Representations}, 2025.
\newblock URL \url{https://openreview.net/forum?id=BC4lIvfSzv}.

\bibitem[Ni et~al.(2022)Ni, Qu, Lu, Dai, Hernandez~Abrego, Ma, Zhao, Luan, Hall, Chang, and Yang]{gtr}
Ni, J., Qu, C., Lu, J., Dai, Z., Hernandez~Abrego, G., Ma, J., Zhao, V., Luan, Y., Hall, K., Chang, M.-W., and Yang, Y.
\newblock Large dual encoders are generalizable retrievers.
\newblock In Goldberg, Y., Kozareva, Z., and Zhang, Y. (eds.), \emph{Proceedings of the 2022 Conference on Empirical Methods in Natural Language Processing}, pp.\  9844--9855, Abu Dhabi, United Arab Emirates, December 2022. Association for Computational Linguistics.
\newblock \doi{10.18653/v1/2022.emnlp-main.669}.
\newblock URL \url{https://aclanthology.org/2022.emnlp-main.669/}.

\bibitem[Paszke et~al.(2019)Paszke, Gross, Massa, Lerer, Bradbury, Chanan, Killeen, Lin, Gimelshein, Antiga, Desmaison, K{\"{o}}pf, Yang, DeVito, Raison, Tejani, Chilamkurthy, Steiner, Fang, Bai, and Chintala]{pytorch}
Paszke, A., Gross, S., Massa, F., Lerer, A., Bradbury, J., Chanan, G., Killeen, T., Lin, Z., Gimelshein, N., Antiga, L., Desmaison, A., K{\"{o}}pf, A., Yang, E.~Z., DeVito, Z., Raison, M., Tejani, A., Chilamkurthy, S., Steiner, B., Fang, L., Bai, J., and Chintala, S.
\newblock {PyTorch}: An imperative style, high-performance deep learning library.
\newblock In Wallach, H.~M., Larochelle, H., Beygelzimer, A., d'Alch{\'{e}}{-}Buc, F., Fox, E.~B., and Garnett, R. (eds.), \emph{Advances in Neural Information Processing Systems 32: Annual Conference on Neural Information Processing Systems 2019, NeurIPS 2019, December 8-14, 2019, Vancouver, BC, Canada}, pp.\  8024--8035, 2019.

\bibitem[Robertson \& Walker(1994)Robertson and Walker]{bm25}
Robertson, S.~E. and Walker, S.
\newblock Some simple effective approximations to the 2-poisson model for probabilistic weighted retrieval.
\newblock In Croft, W.~B. and van Rijsbergen, C.~J. (eds.), \emph{Proceedings of the 17th Annual International {ACM-SIGIR} Conference on Research and Development in Information Retrieval. Dublin, Ireland, 3-6 July 1994 (Special Issue of the {SIGIR} Forum)}, pp.\  232--241. ACM/Springer, 1994.
\newblock \doi{10.1007/978-1-4471-2099-5\_24}.

\bibitem[Roth et~al.(2024)Roth, Udandarao, Dziadzio, Prabhu, Cherti, Vinyals, Henaff, Albanie, Bethge, and Akata]{cont_pretrain_roth_multimodal}
Roth, K., Udandarao, V., Dziadzio, S., Prabhu, A., Cherti, M., Vinyals, O., Henaff, O.~J., Albanie, S., Bethge, M., and Akata, Z.
\newblock A practitioner's guide to continual multimodal pretraining.
\newblock In \emph{NeurIPS 2024 Workshop on Scalable Continual Learning for Lifelong Foundation Models}, 2024.
\newblock URL \url{https://openreview.net/forum?id=gkyosluSbR}.

\bibitem[Sarthi et~al.(2024)Sarthi, Abdullah, Tuli, Khanna, Goldie, and Manning]{raptor}
Sarthi, P., Abdullah, S., Tuli, A., Khanna, S., Goldie, A., and Manning, C.~D.
\newblock {RAPTOR:} recursive abstractive processing for tree-organized retrieval.
\newblock In \emph{The Twelfth International Conference on Learning Representations, {ICLR} 2024, Vienna, Austria, May 7-11, 2024}. OpenReview.net, 2024.
\newblock URL \url{https://openreview.net/forum?id=GN921JHCRw}.

\bibitem[Shi et~al.(2024)Shi, Xu, Wang, Qin, Wang, Wang, Wang, Ebrahimi, and Wang]{shi2024continual}
Shi, H., Xu, Z., Wang, H., Qin, W., Wang, W., Wang, Y., Wang, Z., Ebrahimi, S., and Wang, H.
\newblock Continual learning of large language models: A comprehensive survey.
\newblock \emph{arXiv preprint arXiv:2404.16789}, 2024.

\bibitem[Suzuki(2005)]{suzuki2005associative}
Suzuki, W.~A.
\newblock Associative learning and the hippocampus.
\newblock \emph{Psychological Science Agenda}, February 2005.

\bibitem[Thakur et~al.(2021)Thakur, Reimers, R{\"u}ckl{\'e}, Srivastava, and Gurevych]{beir}
Thakur, N., Reimers, N., R{\"u}ckl{\'e}, A., Srivastava, A., and Gurevych, I.
\newblock {BEIR}: A heterogeneous benchmark for zero-shot evaluation of information retrieval models.
\newblock In \emph{Thirty-fifth Conference on Neural Information Processing Systems Datasets and Benchmarks Track (Round 2)}, 2021.
\newblock URL \url{https://openreview.net/forum?id=wCu6T5xFjeJ}.

\bibitem[Trivedi et~al.(2022)Trivedi, Balasubramanian, Khot, and Sabharwal]{musique}
Trivedi, H., Balasubramanian, N., Khot, T., and Sabharwal, A.
\newblock {MuSiQue}: Multihop questions via single-hop question composition.
\newblock \emph{Transactions of the Association for Computational Linguistics}, 10:\penalty0 539--554, 2022.
\newblock \doi{10.1162/tacl_a_00475}.
\newblock URL \url{https://aclanthology.org/2022.tacl-1.31/}.

\bibitem[Uner \& Roediger~III(2022)Uner and Roediger~III]{Uner2022}
Uner, O. and Roediger~III, H.~L.
\newblock Do recall and recognition lead to different retrieval experiences?
\newblock \emph{The American Journal of Psychology}, 135\penalty0 (1):\penalty0 33--43, 2022.

\bibitem[Wang et~al.(2024)Wang, Ren, Li, Zhao, Liu, and Wen]{rear}
Wang, Y., Ren, R., Li, J., Zhao, X., Liu, J., and Wen, J.
\newblock {REAR:} {A} relevance-aware retrieval-augmented framework for open-domain question answering.
\newblock In Al{-}Onaizan, Y., Bansal, M., and Chen, Y. (eds.), \emph{Proceedings of the 2024 Conference on Empirical Methods in Natural Language Processing, {EMNLP} 2024, Miami, FL, USA, November 12-16, 2024}, pp.\  5613--5626. Association for Computational Linguistics, 2024.
\newblock URL \url{https://aclanthology.org/2024.emnlp-main.321}.

\bibitem[Wolf et~al.(2019)Wolf, Debut, Sanh, Chaumond, Delangue, Moi, Cistac, Rault, Louf, Funtowicz, and Brew]{huggingface}
Wolf, T., Debut, L., Sanh, V., Chaumond, J., Delangue, C., Moi, A., Cistac, P., Rault, T., Louf, R., Funtowicz, M., and Brew, J.
\newblock Huggingface's transformers: State-of-the-art natural language processing.
\newblock \emph{CoRR}, abs/1910.03771, 2019.
\newblock URL \url{http://arxiv.org/abs/1910.03771}.

\bibitem[Xie et~al.(2024)Xie, Zhang, Chen, Lou, and Su]{xie2024adaptive}
Xie, J., Zhang, K., Chen, J., Lou, R., and Su, Y.
\newblock Adaptive chameleon or stubborn sloth: Revealing the behavior of large language models in knowledge conflicts.
\newblock In \emph{The Twelfth International Conference on Learning Representations}, 2024.
\newblock URL \url{https://openreview.net/forum?id=auKAUJZMO6}.

\bibitem[Yao et~al.(2023)Yao, Wang, Tian, Cheng, Li, Deng, Chen, and Zhang]{yao23editing}
Yao, Y., Wang, P., Tian, B., Cheng, S., Li, Z., Deng, S., Chen, H., and Zhang, N.
\newblock Editing large language models: Problems, methods, and opportunities.
\newblock In Bouamor, H., Pino, J., and Bali, K. (eds.), \emph{Proceedings of the 2023 Conference on Empirical Methods in Natural Language Processing}, pp.\  10222--10240, Singapore, December 2023. Association for Computational Linguistics.
\newblock \doi{10.18653/v1/2023.emnlp-main.632}.
\newblock URL \url{https://aclanthology.org/2023.emnlp-main.632/}.

\bibitem[Yuan et~al.(2024)Yuan, Ning, Zhou, Yang, Li, Zhuang, Tan, Yao, Lin, Li, Dai, Yan, and Wang]{lveval}
Yuan, T., Ning, X., Zhou, D., Yang, Z., Li, S., Zhuang, M., Tan, Z., Yao, Z., Lin, D., Li, B., Dai, G., Yan, S., and Wang, Y.
\newblock {LV-Eval}: A balanced long-context benchmark with 5 length levels up to 256k, 2024.
\newblock URL \url{https://arxiv.org/abs/2402.05136}.

\bibitem[Zhang et~al.(2024)Zhang, Gui, Zhai, Wang, Lei, and Xu]{copr}
Zhang, H., Gui, L., Zhai, Y., Wang, H., Lei, Y., and Xu, R.
\newblock Copr: Continual learning human preference through optimal policy regularization, 2024.
\newblock URL \url{https://arxiv.org/abs/2310.15694}.

\bibitem[Zhang et~al.(2023)Zhang, Fang, Chen, and Namazi-Rad]{citb}
Zhang, Z., Fang, M., Chen, L., and Namazi-Rad, M.-R.
\newblock {CITB}: A benchmark for continual instruction tuning.
\newblock In Bouamor, H., Pino, J., and Bali, K. (eds.), \emph{Findings of the Association for Computational Linguistics: EMNLP 2023}, pp.\  9443--9455, Singapore, December 2023. Association for Computational Linguistics.
\newblock \doi{10.18653/v1/2023.findings-emnlp.633}.
\newblock URL \url{https://aclanthology.org/2023.findings-emnlp.633/}.

\bibitem[Zhong et~al.(2023)Zhong, Wu, Manning, Potts, and Chen]{zhong-etal-2023-mquake}
Zhong, Z., Wu, Z., Manning, C., Potts, C., and Chen, D.
\newblock {MQ}u{AKE}: Assessing knowledge editing in language models via multi-hop questions.
\newblock In Bouamor, H., Pino, J., and Bali, K. (eds.), \emph{Proceedings of the 2023 Conference on Empirical Methods in Natural Language Processing}, pp.\  15686--15702, Singapore, December 2023. Association for Computational Linguistics.
\newblock \doi{10.18653/v1/2023.emnlp-main.971}.
\newblock URL \url{https://aclanthology.org/2023.emnlp-main.971/}.

\end{thebibliography}
\bibliographystyle{icml2025}

%%%%%%%%%%%%%%%%%%%%%%%%%%%%%%%%%%%%%%%%%%%%%%%%%%%%%%%%%%%%%%%%%%%%%%%%%%%%%%%
%%%%%%%%%%%%%%%%%%%%%%%%%%%%%%%%%%%%%%%%%%%%%%%%%%%%%%%%%%%%%%%%%%%%%%%%%%%%%%%
% APPENDIX
%%%%%%%%%%%%%%%%%%%%%%%%%%%%%%%%%%%%%%%%%%%%%%%%%%%%%%%%%%%%%%%%%%%%%%%%%%%%%%%
%%%%%%%%%%%%%%%%%%%%%%%%%%%%%%%%%%%%%%%%%%%%%%%%%%%%%%%%%%%%%%%%%%%%%%%%%%%%%%%
\newpage
\appendix
\onecolumn

\section*{Appendices}

Within this supplementary material, we elaborate on the following aspects:

\begin{itemize}
    \item Appendix \ref{sec:prompts}: LLM Prompts
    \item Appendix \ref{sec:pipeline example}: \ours Pipeline Example
    \item Appendix \ref{sec:detailed results}: Detailed Experimental Results
    \item Appendix \ref{sec:graph statistics}: Graph Statistics
    \item Appendix \ref{sec:error analysis}: Error Analysis
    \item Appendix \ref{sec:cost and efficiency}: Cost and Efficiency
    \item Appendix \ref{sec:appendix-hyper}: Implementation Details and Hyperparameters
\end{itemize}

\section{LLM Prompts}
\label{sec:prompts}

\begin{figure*}
    \centering
    \includegraphics[width=0.85\linewidth]{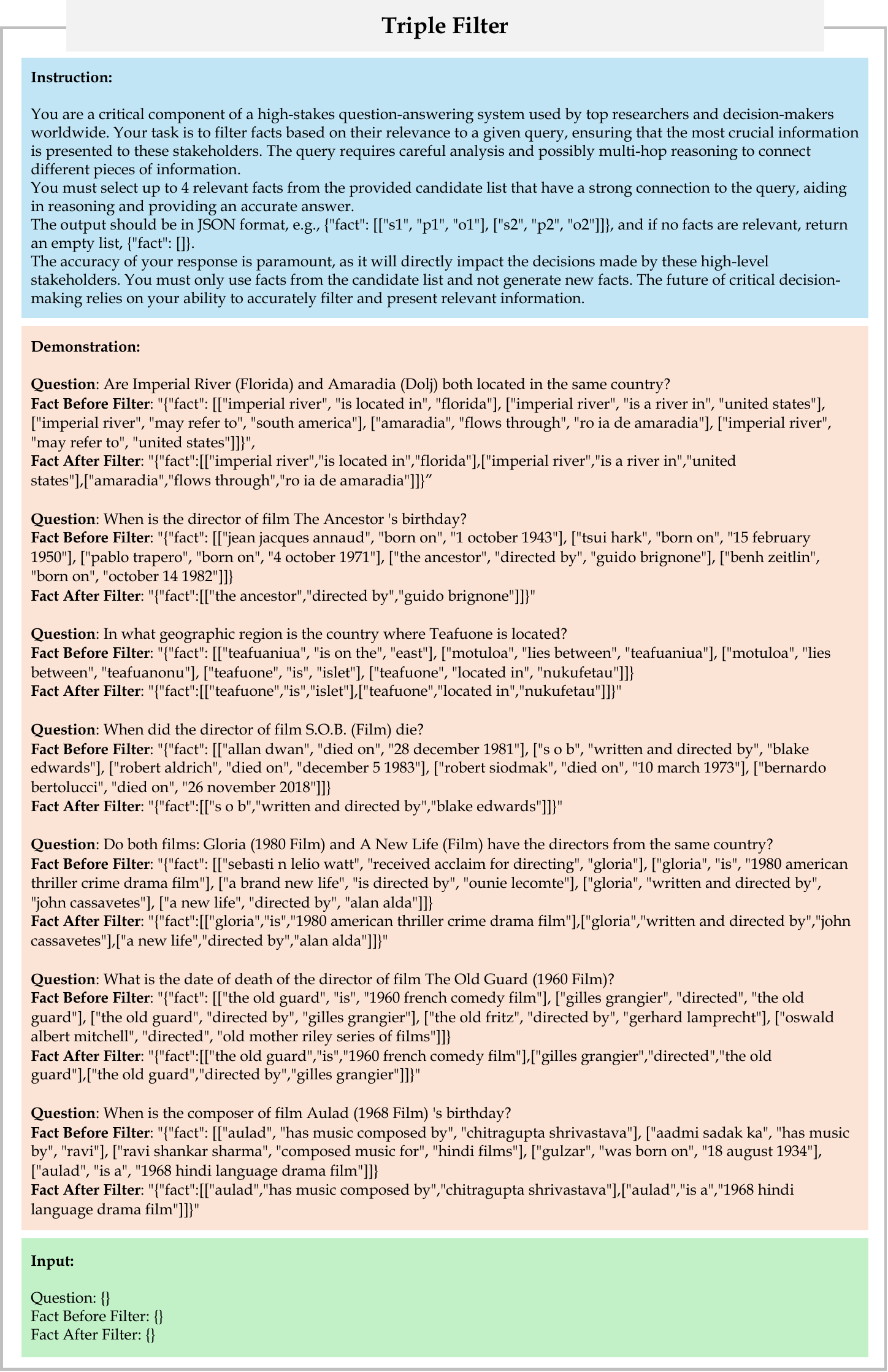}
    \caption{LLM prompts for triple filtering (recognition memory).}
    \label{fig:prompts for triple filtering}
\end{figure*}

We show LLM prompts for triple filter in Figure \ref{fig:prompts for triple filtering}, including the instruction, the few-shot demonstrations and the input format.

\section{Pipeline Example}
\label{sec:pipeline example}

\begin{figure*}
    \centering
    \includegraphics[width=0.73\linewidth]{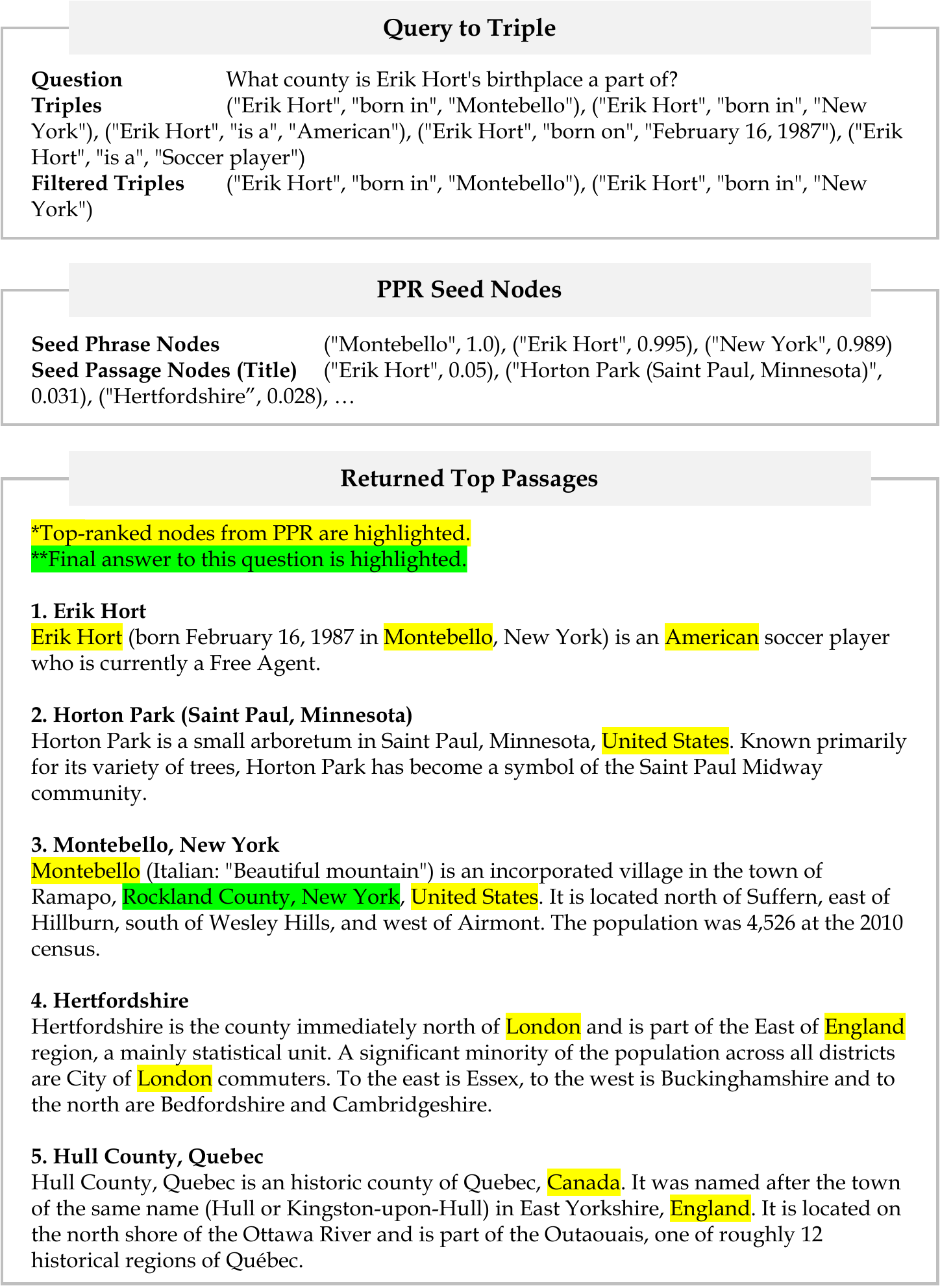}
    \caption{An example of \ours pipeline.}
    \label{fig:pipeline example}
\end{figure*}

We show a pipeline example of \ours online retrieval in Figure \ref{fig:pipeline example}, including query-to-triple, triple filtering and using seed nodes for PPR.

\section{Detailed Experimental Results}
\label{sec:detailed results}

We show QA performance and retrieval performance with the proprietary model GPT-4o-mini as well as more metrics here, as shown in Table \ref{table:appendix qa performance} and Table \ref{table:appendix recall results}.

\noindent \textbf{QA Performance}
As shown in Table \ref{table:appendix qa performance}, when using GPT-4o-mini for indexing and QA reading, \ours consistently achieves competitive EM and F1 scores across most datasets. Notably, it leads in the MuSiQue and 2Wiki benchmarks. Our method also demonstrates superior performance in the NarrativeQA and LV-Eval tasks. When compared to the strong NV-Embed-v2 retriever, \ours exhibits comparable or enhanced F1 scores, particularly excelling in the LV-Eval dataset with reduced knowledge leakage.

\noindent \textbf{Retrieval Performance}
As shown in Table \ref{table:appendix recall results}, the improvement trend of \ours in recall@2 is similar to that in recall@5.

\begin{table*}[htb]
  \centering
  \small
    \caption{QA performance (EM / F1 scores) on RAG benchmarks. No retrieval means evaluating the parametric knowledge of the readers. HippoRAG (and \ours) uses the denoted LLM for OpenIE (triple filtering) and QA reading.}
    \vskip 0.1in
  \resizebox{\textwidth}{!}{%
  \begin{tabular}{lccccccccccccc}
  \toprule
& \multicolumn{2}{c}{\multirow{2}{*}{Simple QA}} 
& \multicolumn{4}{c}{\multirow{2}{*}{Multi-Hop QA}} 
& \multicolumn{1}{c}{\multirow{2}{*}{\shortstack{Discourse\\Understanding}}} & \\
&&&&&&&&\\
  \cmidrule(lr){2-3} \cmidrule(lr){4-7} \cmidrule(lr){8-8}
  Retrieval & NQ & PopQA & MuSiQue & 2Wiki & HotpotQA & LV-Eval & NarrativeQA & Avg \\
   \midrule
   \multicolumn{9}{c}{Llama-3.3-70B-Instruct} \\
   \midrule
% \rowcolor{lightgrey}\multicolumn{9}{c}{\textit{\textbf{Simple Baselines}}} \\
  None & $40.2$ / $54.9$ & $28.2$ / $32.5$ & $17.6$ / $26.1$ & $36.5$ / $42.8$ & $37.0$ / $47.3$ & \ \ $4.0$ / \ \ $6.0$ &  \ \ $3.4$ / $12.9$ & $29.7$ / $38.4$ \\ 
  Contriever~\cite{contriever} & $45.0$ / $58.9$ & $41.6$ / $53.1$ & $24.0$ / $31.3$ & $38.1$ / $41.9$ & $51.3$ / $62.3$ & \ \ $5.7$ / \ \ $8.1$ & \ \ $6.5$ / $19.7$ & $37.4$ / $46.9$ \\
  BM25~\cite{bm25} & $44.7$ / $59.0$ & $39.1$ / $49.9$ & $20.3$ / $28.8$ & $47.9$ / $51.2$ & $52.0$ / $63.4$ & \ \ $4.0$ / \ \ $5.9$ & \ \ $4.4$ / $18.3$ & $38.0$ / $47.7$ \\ 
% \rowcolor{lightgrey}\multicolumn{9}{c}{\textit{\textbf{Large Embedding Models}}} \\
  GTR (T5-base)~\cite{gtr} & $45.5$ / $59.9$ & $\underline{43.2}$ / $\underline{56.2}$  & $25.8$ / $34.6$ & $49.2$ / $52.8$ & $50.6$ / $62.8$ & \ \ $4.8$ / \ \ $7.1$ & \ \ $6.8$ / $19.9$ & $40.0$ / $50.4$ \\
  GTE-Qwen2-7B-Instruct~\cite{gte} & $46.6$ / $\underline{62.0}$ & $\mathbf{43.5}$ / $\mathbf{56.3}$ & $30.6$ / $40.9$ & $55.1$ / $60.0$ & $58.6$ / $71.0$ & \ \ $5.7$ / \ \ $7.1$ & \ \ $7.9$ / $21.3$ & $43.8$ / $54.9$ \\
  GritLM-7B~\cite{gritlm} & $46.8$ / $61.3$ & $42.8$ / $55.8$ & $33.6$ / $44.8$ & $55.8$ / $60.6$ & $60.7$ / $73.3$ & \ \ $\underline{7.3}$ / \ \ $9.8$ & \ \ $\underline{8.2}$ / $23.9$ & $44.9$ / $56.1$ \\
  NV-Embed-v2 (7B)~\cite{nvembedv2} &  $\underline{47.3}$ / $61.9$	& $42.9$ / $55.7$ & $\underline{34.7}$ / $\underline{45.7}$ & $\underline{57.5}$ / $61.5$ &	$\mathbf{62.8}$ / $\underline{75.3}$ & \ \ $\underline{7.3}$ / \ \ $9.8$  & \ \ $\mathbf{8.9}$ / $\underline{25.7}$ & $\underline{45.9}$ / $\underline{57.0}$ \\
  % \rowcolor{lightgrey}\multicolumn{9}{c}{\textit{\textbf{Structure-augmented RAG}}} \\
  RAPTOR~\cite{raptor} & $36.9$ / $50.7$& $43.1$ / $\underline{56.2}$ & $20.7$ / $28.9$ & $47.3$ / $52.1$  & $56.8$ / $69.5$  & \ \ $2.4$ / \ \ $5.0$ & \ \ $5.1$ / $21.4$ & $38.1$ / $48.8$ \\
  GraphRAG~\cite{graphrag} & $30.8$ / $46.9$ & $31.4$ / $48.1$ &$27.3$ / $38.5$  &  $51.4$ / $58.6$  & $55.2$ / $68.6$ & \ \ $4.8$ / $\underline{11.2}$ & \ \ $6.8$ / $23.0$ & $36.7$ / $49.6$ \\
  LightRAG~\cite{lightrag} & \ \ $8.6$ / $16.6$  & \ \ $2.1$ / \ \ $2.4$ & \ \ $0.5$ / \ \ $1.6$ & \ \ $9.4$ / $11.6$  & \ \ $2.0$ / \ \ $2.4$ & \ \ $0.8$ / \ \ $1.0$ & \ \ $1.0$ / \ \ $3.7$ & \ \ $4.2$ / \ \ $6.6$ \\
  HippoRAG~\cite{hipporag} & $43.0$ / $55.3$ & $42.7$ / $55.9$ & $26.2$ / $35.1$ &	$\mathbf{65.0}$ / $\mathbf{71.8}$ &	$52.6$ / $63.5$  & \ \ $6.5$ / \ \ $8.4$ & \ \ $4.4$ / $16.3$ & $42.8$ / $53.1$ \\
  \ours  & $\mathbf{48.6}$ / $\mathbf{63.3}$ & $42.9$ / $\underline{56.2}$ & $\mathbf{37.2}$ / $\mathbf{48.6}$ & $\mathbf{65.0}$ / $\underline{71.0}$ & $\underline{62.7}$ / $\mathbf{75.5}$ & \ \ $\mathbf{9.7}$ / $\mathbf{12.9}$ & \ \ $\mathbf{8.9}$ / $\mathbf{25.9}$ & $\mathbf{48.0}$ /	$\mathbf{59.8}$ \\
  \midrule
     \multicolumn{9}{c}{GPT-4o-mini} \\ 
     \midrule
     % \rowcolor{lightgrey}\multicolumn{9}{c}{\textit{\textbf{Simple Baselines}}} \\
  None & $35.2$ / $52.7$ & $16.1$ / $22.7$ & $11.2$ / $22.0$ &  $30.2$ / $36.3$ & $28.6$ / $41.0$ & \ \ $3.2$ / \ \ $5.0$ & \ \ $2.7$ / $14.1$  & $22.6$ / $33.1$ \\
  % \rowcolor{lightgrey}\multicolumn{9}{c}{\textit{\textbf{Large Embedding Models}}} \\
  NV-Embed-v2 (7B)~\cite{nvembedv2} & $\mathbf{43.5}$ / $\underline{59.9}$ & $41.7$ / $\underline{55.8}$ & $\underline{32.8}$ / $\underline{46.0}$ & $54.4$ / $60.8$ & $\mathbf{57.3}$ / $\underline{71.0}$ &  \ \ $\underline{7.3}$ / $10.0$ & \ \ $5.1$ / $\underline{24.2}$ & $\underline{42.9}$ / $\underline{55.7}$ \\
  % \midrule
  % \rowcolor{lightgrey}\multicolumn{9}{c}{\textit{\textbf{Structure-augmented RAG}}} \\
  RAPTOR~\cite{raptor} & $37.8$ / $54.5$& $\underline{41.9}$ / $55.1$ & $27.7$ / $39.2$ & $39.7$ / $48.4$ & $50.6$ / $64.7$  & \ \ $5.6$ / \ \ $9.2$ & \ \ $4.1$ / $21.8$ & $36.9$ / $49.7$ \\
  GraphRAG~\cite{graphrag} & $38.0$ / $55.5$ & $30.7$ / $51.3$ & $27.0$ / $42.0$  & $45.7$ /\ $61.0$  & $51.4$ / $67.6$ & \ \ $4.9$ / $\underline{11.0}$  & \ \ $\underline{5.4}$ / $20.9$ & $36.0$ / $52.6$  \\
  LightRAG~\cite{lightrag} & \ \ $2.8$ / $15.4$ & \ \ $1.9$ / $14.8$& \ \ $2.0$ / \ \ $9.3$ & \ \ $2.5$ /\ $12.1$ & \ \ $9.9$ /\ $20.2$ & \ \ $0.9$ / \ \ $5.0$ & \ \ $1.0$ / \ \ $9.0$ & \ \ $3.6$ / $13.9$ \\
  HippoRAG~\cite{hipporag} & $37.2$ / $52.2$ & $\mathbf{42.5}$ / $\mathbf{56.2}$ & $24.0$ / $35.9$ & $\underline{59.4}$ / $\underline{67.3}$ & $46.3$ / $60.0$ & \ \ $4.8$ / \ \ $7.6$ & \ \ $2.1$ / $16.1$ & $38.9$ / $51.2$ \\ 
  % \midrule
  \ours & $\underline{43.4}$ / $\mathbf{60.0}$ & $41.7$ / $55.7$ & $\mathbf{35.0}$ / $\mathbf{49.3}$ & $\mathbf{60.5}$ / $\mathbf{69.7}$ & $\underline{56.3}$ / $\mathbf{71.1}$ & $\mathbf{10.5}$ / $\mathbf{14.0}$ & \ \ $\mathbf{5.8}$ / $\mathbf{25.2}$ & $\mathbf{44.3}$ / $\mathbf{58.1}$ \\
  \bottomrule
  \end{tabular}
  }
  \label{table:appendix qa performance}
  % \vspace{-10pt}
    \vskip -0.1in
\end{table*}

\begin{table*}[htb]
  \centering
  \small
      \caption{Passage recall@2 / @5 on RAG benchmarks. * denotes the report from the original paper while we reproduce the HippoRAG results with aligned LLM and retriever.}
          \vskip 0.1in
  \resizebox{\textwidth}{!}{%
  \begin{tabular}{lcccccc}
  \toprule
  & \multicolumn{2}{c}{Simple}  & \multicolumn{3}{c}{Multi-hop} \\
  \cmidrule(lr){2-3} \cmidrule(lr){4-6}
   & NQ & PopQA & MuSiQue & 2Wiki & HotpotQA & Avg \\
  \midrule
\rowcolor{lightgrey}
  \multicolumn{7}{c}{\textit{\textbf{Simple Baselines}}} \\
  Contriever~\cite{contriever} & $29.1$ / $54.6$ & $27.0$ / $43.2$ & $34.8$ / $46.6$ & $46.6$ / $57.5$ & $58.4$ / $75.3$ & $39.2$ / $55.4$ \\
  BM25~\cite{bm25} & $28.2$ / $56.1$ & $24.0$ / $35.7$ & $32.4$ / $43.5$ & $55.3$ / $65.3$ & $57.3$ / $74.8$  & $39.4$ / $55.1$ \\ 
  % \midrule
  GTR (T5-base)~\cite{gtr} & $35.0$ / $63.4$ & $40.1$ / $49.4$ & $37.4$ / $49.1$ & $60.2$ / $67.9$ & $59.3$ / $73.9$ & $46.4$ / $60.7$ \\
  \rowcolor{lightgrey}
  \multicolumn{7}{c}{\textit{\textbf{Large Embedding Models}}} \\
  GTE-Qwen2-7B-Instruct~\cite{gte} & $44.7$ / $74.3$ & $\mathbf{47.7}$ / $50.6$ & $48.1$ / $63.6$ & $66.7$ / $74.8$ & $75.8$ / $89.1$ & $56.6$ / $70.5$  \\
  GritLM-7B~\cite{gritlm} & $\mathbf{46.2}$ / $\underline{76.6}$ & $44.0$ / $50.1$ & $49.7$ / $65.9$ & $67.3$ / $76.0$ & $79.2$ / $92.4$ & $57.3$ / $72.2$ \\
  NV-Embed-v2 (7B)~\cite{nvembedv2} & $45.3$ / $75.4$ & $\underline{45.3}$ / $51.0$ & $52.7$ / $69.7$ & $67.1$ / $76.5$ & $\mathbf{84.1}$ / $94.5$ & $58.9$ / $73.4$ \\ 
  \rowcolor{lightgrey}
  \multicolumn{7}{c}{\textit{\textbf{Structure-augmented RAG}}} \\
  % RAPTOR~\cite{raptor} &$40.5$ / $69.4$ & $37.2$ / $48.1$ & $49.1$ / $61.0$ & $61.5$ / $70.6$ & $78.6$ / $90.2$  \\
  RAPTOR~(GPT-4o-mini) & $40.5$ / $69.4$  & $37.2$ / $48.1$ &  $49.1$ / $61.0$ & $58.4$ / $66.0$ & $78.6$ / $90.2$ & $52.8$ / $67.0$  \\
  RAPTOR~(Llama-3.3-70B-Instruct) & $40.3$ / $68.3$  & $40.2$ / $48.7$   &  $47.0$ / $57.8$ & $58.3$ / $66.2$ & $76.8$ / $86.9$ & $52.5$ / $65.6$  \\
  % GraphRAG & & & & & & & \\
  % LightRAG & & & & & & & \\
  HippoRAG*~\cite{hipporag} & $-$ & $-$ & $40.9$ / $51.9$ & $70.7$ / $89.1$ & 
 $60.5$ / $77.7$ & $-$ \\
  HippoRAG~(GPT-4o-mini) & $21.6$ / $45.1$ & $36.5$ / $52.2$ & $41.8$ / $52.4$ & $68.4$ / $87.0$ & $60.1$ / $78.5$ & $45.7$ / $63.0$ \\ 
   HippoRAG~(Llama-3.3-70B-Instruct) & $21.3$ / $44.4$ & $40.0$ / $\mathbf{53.8}$ & $41.2$ / $53.2$ & $71.9$ / $\mathbf{90.4}$ & $60.4$ / $77.3$ & $47.0$ / $63.8$ \\
   % KAG*~\cite{} & - & - & 48.5 / 65.7 & 65.4 / 91.9 & - \\
   \midrule
  \ours (GPT-4o-mini) & $44.4$ / $76.4$ & $43.5$ / $\underline{52.2}$ &  $\underline{53.5}$ / $\underline{74.2}$ & $\underline{74.6}$ / $\underline{90.2}$ & $80.5$ / $\underline{95.7}$ & $\underline{59.3}$ / $\underline{77.7}$  \\
  \ours (Llama-3.3-70B-Instruct) & $\underline{45.6}$ / $\mathbf{78.0}$ & $43.9$ / $51.7$ & $\mathbf{56.1}$ / $\mathbf{74.7}$ & $\mathbf{76.2}$ / $\mathbf{90.4}$ & $\underline{83.5}$ / $\mathbf{96.3}$ & $\mathbf{61.1}$ / $\mathbf{78.2}$ \\
  % \ours & & & $\mathbf{55.4}$ / $\mathbf{73.3}$ & $\mathbf{74.1}$ / $\mathbf{87.4}$ & $\mathbf{84.9}$ / $\mathbf{95.9}$ % 4o filter\\
  \bottomrule
  \end{tabular}
  }
  % \vspace{-10pt}

  \label{table:appendix recall results}
    \vskip -0.1in
\end{table*}

\section{Graph Statistics}
\label{sec:graph statistics}

We show the knowledge graph statistics using Llama-3.3-70B-Instruct or GPT-4o-mini for OpenIE in Table \ref{tab:graph statistics}.

\begin{table}[]
    \centering
    \small
        \caption{Knowledge graph statistics using different LLMs for OpenIE. The nodes and triples are counted based on unique values. }
        \vskip 0.1in
    \resizebox{\textwidth}{!}{%
    \begin{tabular}{lrrrrrrr}
    \toprule
     & NQ & PopQA & MuSiQue & 2Wiki & HotpotQA & LV-Eval & NarrativeQA \\ 
     \midrule
     \multicolumn{8}{c}{Llama-3.3-70B-Instruct} \\
     \midrule
        \# of phrase nodes & $68,375$ & $76,539$ & $85,288$ & $44,004$ & $81,200$ & $175,195$ & 
$9,224$ \\
        \# of passage nodes & $9,633$ & $8,676$ & $11,656$ & $6,119$ & $9,811$ & $22,849$ & $4,111$ \\
        \# of total nodes & $78,008$ & $85,215$ & $96,944$ & $50,123$ & $91,011$ & $198,044$ & $13,335$ \\ 
        \# of extracted edges & $125,777$ & $124,579$ & $140,830$ & $68,881$ & $130,058$ & $314,324$ & $26,208$\\
        \# of synonym edges & $899,031$ & $845,014$ & $1,125,951$ & $593,298$ & $994,187$ & $2,674,833$ & $72,494$ \\
        \# of context edges & $126,757$ & $118,909$ & $132,586$ & $64,132$ & $122,437$ & $375,424$ & $33,395$ \\
        \# of total edges & $1,151,565$ & $1,088,502$ & $1,399,367$ & $726,311$ & $1,246,682$ & $3,364,581$ & $132,097$ \\ \midrule
    \multicolumn{8}{c}{GPT-4o-mini} \\ \midrule
           \# of phrase nodes & $86,904$ & $85,744$ & $101,641$ & $49,544$ & $95,105$ & $217,085$ & $15,365$ \\
        \# of passage nodes & $9,633$ & $8,676$ & $11,656$ & $6,119$ & $9,811$ & $22,849$ & $4,111$ \\
        \# of total nodes & $96,537$ & $94,420$ & $113,297$ & $55,663$ & $104,916$ & $239,934$ & $19,476$ \\ 
        \# of extracted edges & $114,900$ & $108,989$ & $125,903$ & $62,626$ & $119,630$ & $303,491$ & $24,373$ \\
        \# of synonym edges & $1,094,651$ & $901,528$ & $1,304,605$ & $715,763$ & $1,126,501$ & $3,268,084$ & $14,075$ \\
        \# of context edges & $142,419$ & $127,568$ & $146,293$ & $68,348$ & $133,220$ & $404,210$ & $38,632$ \\
        \# of total edges & $1,351,970$ & $494,082$ & $1,576,801$ & $846,737$ & $1,379,351$ & $3,975,785$ & $77,080$ \\ 
        \bottomrule
         & 
    \end{tabular}
    }
    \label{tab:graph statistics}
    \vskip -0.1in
\end{table}

\section{Error Analysis}
\label{sec:error analysis}

We provide an error analysis of 100 samples generated by \ours with recall@5 less than 1.0. 
Among these samples, 26\%, 41\%, and 33\% are classified as 2-hop, 3-hop, and 4-hop questions, respectively. 
Triple filtering and the graph search algorithm are the two main sources of errors.

\noindent \textbf{Recognition Memory} In 7\% of the samples, no phrase from the supporting documents is matched with the phrases obtained by the query-to-triple stage before triple filtering. 
In 26\% of the samples, no phrase from the supporting documents is matched with the phrases after triple filtering.
After the triple filtering step, 8\% of the samples show a decrease in the proportion of phrases in the triples that match phrases from the supporting passages. For instance, the first case from Table \ref{tab:error examples} shows an empty list after triple filtering, which eliminates all relevant phrases.
Additionally, 18\% of the samples are left with zero triples after filtering. Although not necessarily an error in filtering, this indicates that the attempt to link to the triples has failed, where \ours directly uses the results from dense retrieval as a substitute. Overall, though recognition memory is an essential component, the precision of the triple filter has room for further improvement.

\noindent \textbf{Graph Construction} Graph construction is challenging to evaluate, but we find that only 2\% of the samples do not contain any phrases from the supporting passages within the one-hop neighbors of the linked nodes. Given our dense-sparse integration, we can assume that the graphs we construct generally include most of the potentially exploitable information.

\noindent \textbf{Personalized PageRank} In 50\% of the samples, at least half of the linked phrase nodes appear in the supporting documents. However, the final results remain unsatisfactory due to the graph search component. For example, in the second case from Table \ref{tab:error examples}, the recognition memory identifies the key phrase "Philippe, Duke of Orléans" from the query, but the graph search fails to return perfect results among the top-5 retrieved passages.

\begin{table*}[]
\small
\centering
\caption{Two examples from MuSiQue where passage recall@5 is less than 1.0.}
    \vskip 0.1in
\resizebox{\columnwidth}{!}{%
\begin{tabular}{lp{13cm}}
\toprule
\textbf{Query} & Where is the district that the person who wanted to reform and address Bernhard Lichtenberg's religion preached a sermon on Marian devotion before his death located? \\ 
\textbf{Answer} & Saxony-Anhalt \\ \midrule
\textbf{Supporting Passages (Title)} & 1. Mary, mother of Jesus 2. Reformation 3. Wittenberg (district)  4. Bernhard Lichtenberg \\
\textbf{Retrieved Passages (Title)} & \textbf{1. Bernhard Lichtenberg} \textbf{2. Mary, mother of Jesus} 3. Ambroise-Marie Carré \textbf{4. Reformation} 5. Henry Scott Holland (Recall@5 is 0.75) \\ \midrule
\textbf{Query to Triple (Top-5)} & \begin{tabular}[c]{@{}l@{}}("Bernhard Lichtenberg", "was", "Roman Catholic Priest")\\ ("Bernhard Lichtenberg", "beatified by", "Catholic Church")\\ ("Bernhard Lichtenberg", "died on", "5 November 1943")\\ ("Catholic Church", "beatified", "Bernhard Lichtenberg")\\ ("Bernhard Lichtenberg", "was", "Theologian")\\ All above subjects and objects appear in supporting passages\end{tabular} \\ \midrule
\textbf{Filtered Triple} & Empty \\ 

\bottomrule
\toprule

\textbf{Query} & Who is the grandmother of Philippe, Duke of Orléans? \\
\textbf{Answer} & Marie de' Medici \\ \midrule
\textbf{Supporting Passages (Title)} & 1. Philippe I, Duke of Orléans 2. Leonora Dori \\
\textbf{Retrieved Passages (Title)} & \textbf{1. Philippe I, Duke of Orléans} 2. Louise Élisabeth d'Orléans 3. Philip III of Spain 4. Anna of Lorraine 5. Louis Philippe I (Recall@5 is 0.5)\\ \midrule
\textbf{Query to Triple (Top-5)} & \begin{tabular}[c]{@{}l@{}}("Bank of America", "purchased", "Fleetboston Financial")\\ ("Fleetboston Financial", "was acquired by", "Bank of America")\\ ("Bank of America", "acquired", "Fleetboston Financial")\\ ("Bank of America", "announced purchase of", "Fleetboston Financial") \\ ("Bank of America", "merged with", "Fleetboston Financial")\\ All above subjects and objects appear in supporting passages\end{tabular} \\ \midrule
\textbf{Filtered Triple} & \begin{tabular}[c]{@{}l@{}}("Bank of America", "purchased", "Fleetboston Financial")\\ ("Fleetboston Financial", "was acquired by", "Bank of America")\\  All above subjects and objects appear in supporting passages\end{tabular} \\ 
\bottomrule
\end{tabular}
}
\label{tab:error examples}
    \vskip -0.1in
\end{table*}

\section{Cost and Efficiency}
\label{sec:cost and efficiency}

For LLM deployment, we run Llama-3.3-70B-Instruct on a machine equipped with four NVIDIA H100 GPUs, utilizing tensor parallelism via vLLM \cite{vllm}.

For a detailed comparison with our baselines, we track the computational resources (\# of tokens, indexing time, time per query and GPU memory) usage when indexing and performing QA on the MuSiQue corpus (11k documents) using the Llama-3.3-70B-Instruct model. We compare \ours against NV-Embed-v2 \cite{nvembedv2}, RAPTOR \cite{raptor}, LightRAG \cite{lightrag}, HippoRAG \cite{hipporag} and GraphRAG \cite{graphrag} in Table \ref{tab:token_stats}. For the memory requirements, we ignore all memory for model weights since it is shared across all systems.

\ours not only outperforms these RAG methods in QA and retrieval performance but also uses much fewer tokens compared to LightRAG and GraphRAG. In terms of time, we note that \ours is much more efficient than GraphRAG and LightRAG while only being slightly less efficient than both RAPTOR and HippoRAG. \ours’s use of fact embeddings does increase its memory requirements compared to our baselines, however, we believe that this is an acceptable tradeoff given our method's performance benefits. Additionally, while all approaches lag behind standard RAG in terms of time and memory efficiency, \ours is the only one that outperforms this strong baseline substantially.

\begin{table}[ht]
\small
\centering
\caption{We report the computational resource requirements (indexing tokens, indexing time, time per query, GPU memory requirements during QA) for the RAG baselines on the MuSiQue corpus ($11{,}656$ passages). For each metric, we include the percentage that this method obtains with respect to the HippoRAG 2 metric (100\%).}
\vskip 0.1in
\label{tab:token_stats}
\resizebox{\columnwidth}{!}{%
\begin{tabular}{lcccccc}
\toprule
    & \textbf{NV-Embed-v2} & \textbf{RAPTOR} & \textbf{LightRAG} & \textbf{GraphRAG} & \textbf{HippoRAG} & \textbf{HippoRAG 2} \\
    \midrule
    \textbf{Input Tokens} & -- & $1.7\mathrm{M}$ $(18.5\%)$ & $68.5\mathrm{M}$ $(744.6\%)$ & $115.5\mathrm{M}$ $(1255.4\%)$ & $9.2\mathrm{M}$ $(100.0\%)$ & $9.2\mathrm{M}$ $(100.0\%)$ \\
    \textbf{Output Tokens} & -- & $0.2\mathrm{M}$ $(6.7\%)$  & $18.3\mathrm{M}$ $(610.0\%)$ & $36.1\mathrm{M}$ $(1203.3\%)$ & $3.0\mathrm{M}$ $(100.0\%)$ & $3.0\mathrm{M}$ $(100.0\%)$ \\\midrule
\textbf{Indexing Time (min)} & $12.1$ $(12.3\%)$ & $100.5$ $(101.0\%)$ & $235.0$ $(236.2\%)$ & $277.0$ $(278.4\%)$ & $57.5$ $(57.7\%)$ & $99.5$ $(100.0\%)$  \\
\textbf{QA Time/Query (sec)} & $0.3$ $(25.0\%)$ & $0.6$ $(50.0\%)$ & $13.3$ $(1008.3\%)$ & $10.7$ $(891.7\%)$ & $0.9$ $(75.0\%)$ & $1.2$ $(100.0\%)$\\\midrule
\textbf{QA GPU Memory (GB)} & $1.7$ $(17.2\%)$& $1.4$ $(14.1\%)$& $4.5$ $(45.5\%)$& $3.7$ $(37.4\%)$& $6.0$ $(60.6\%)$& $9.9$ $(100.0\%)$\\
    \bottomrule
\end{tabular}
}
\end{table}
% Ours: OPENIE 3385456, 600569; TRIPLES 5781526, 2371459
% lightrag total input: 68464264, total output: 18269261
% GraphRAG: 115486094, 36074863
% Raptor(1710270, 170783)

\section{Implementation Details and Hyperparameters}
\label{sec:appendix-hyper}

\subsection{\ours}
\label{ours_impl}

% and implement the personalized Pagerank with the PRPACK library\footnote{\url{}}. 

% \paragraph{Personalized PageRank}
\label{PPR_init}

% For the initialization of the PPR process, we use both phrase nodes and passage nodes as seed nodes.
% Phrase nodes undergo a two-stage process to initialize their scores. 
% In the first stage, we select the top-$5$ most relevant triples associated with each phrase node as input to our filtering mechanism. If the number of phrase nodes within the filtered triples exceeds a threshold of $5$, we compute an average node score based on the initial five triple scores. This average score is then used to initialize the scores of these phrase nodes ($N_{\text{entity}}\leq 5$), while other phrase nodes are set to $0$.
% For passage similarity, each passage node is initialized using the cosine similarity of its normalized embedding, since we find that activating a broader set of potential passages is more effective for uncovering passages along multi-hop reasoning chains, compared to focusing only on the top-ranked passages.

We provide a detailed explanation of the PPR initialization process used in \ours here. The key goal is to determine the seed nodes for the PPR search and assign appropriate reset probabilities to ensure an effective retrieval process.

\paragraph{Seed Node Selection} The seed nodes for the PPR search are categorized into two types: phrase nodes and passage nodes. All the scores given by the embedding model below use normalized embedding to calculate.
1) Phrase Nodes: These seed nodes are selected from the phrase nodes within the filtered triples, which are obtained through the recognition memory component. If recognition memory gives an empty triple list and no phrase node is available, \ours directly returns top passages using the embedding model without any graph search. 
Otherwise, we keep at most $5$ phrase nodes as the seed nodes, and the ranking score of each phrase node is computed as the average score of all filtered triples it appears in.
2) Passage Nodes: Each passage node is initially scored using an embedding-based similarity, and these scores are processed as follows. All passage nodes are taken as seed nodes since we find that activating a broader set of potential passages is more effective for uncovering passages along multi-hop reasoning chains compared to focusing only on the top-ranked passages.

\paragraph{Reset Probability Assignment}
After determining the seed nodes, we assign reset probabilities to control how likely the PPR algorithm will return to these nodes during the random walk. The rules are: 1) Phrase nodes receive reset probabilities directly as their ranking scores. 2) Passage nodes receive reset probabilities proportional to their embedding similarity scores, i.e., to balance the influence of phrase nodes and passage nodes, we apply a weight factor to the passage node scores. 
Specifically, the passage node scores are multiplied by the weight factor discussed in Section \ref{subsec:hyperparameter}. This ensures that passage nodes and phrase nodes contribute appropriately to the retrieval process.

\paragraph{PPR Execution and Passage Ranking}
Once the seed nodes and their reset probabilities are initialized, we run PPR over the constructed graph. The final ranking of passages is determined based on the PageRank scores of the passage nodes. Top-ranked passages are then used as inputs for the downstream QA reading process.
We manage our KG and run the PPR algorithm using the python-igraph library.\footnote{\url{https://python.igraph.org/en/stable/}}

By incorporating both phrase nodes and passage nodes into the PPR initialization, our approach ensures a more effective retrieval of relevant passages, especially for multi-hop reasoning tasks.

\paragraph{Hyperparameters} We perform hyperparameter tuning on $100$ examples from MuSiQue’s training data. 
The hyperparameters are listed in Table \ref{hp_ours}.

\begin{table}[h]
\centering
\caption{Hyperparameters set on \ours}
    \vskip 0.1in
\label{hp_ours}
\small
\begin{tabular}{cc}
\toprule
\textbf{Hyperparameter} & \textbf{Value} \\
\midrule
Synonym Threshold & $0.8$ \\
Damping Factor of PPR & $0.5$ \\
Temperature & $0.0$ \\
\bottomrule
\end{tabular}
    \vskip -0.1in
\end{table}

\subsection{Comparison Methods}

We use PyTorch \cite{pytorch} and HuggingFace \cite{huggingface} for dense retrievers and BM25s \cite{bm25s} for the BM25 implementation.
For GraphRAG \cite{graphrag} and LightRAG \cite{lightrag}, we adhere to their default hyperparameters and prompts. 
To ensure a consistent evaluation, the same QA prompt that \ours adopts from HippoRAG \cite{hipporag} is applied to rephrase the original response of GraphRAG and LightRAG. 

\paragraph{Hyperparameters}

We keep the default indexing hyperparameters for GraphRAG and LightRAG. 
For QA, we perform hyperparameter tuning on the same $100$ samples as Appendix \ref{ours_impl}.

\begin{table}[h]
\centering
\caption{Hyperparameters set on GraphRAG and LightRAG}
    \vskip 0.1in
% \resizebox{0.47\textwidth}{!}{%
\small
\begin{tabular}{lrr}
\toprule
\textbf{Hyperparameters} & \textbf{GraphRAG} & \textbf{LightRAG} \\
\midrule
Mode & Local & Local \\
Response Type & Short phrase & Short phrase \\
Top-k Phrases for QA & $60$ & $60$ \\
Chunk Token Size & $1,200$ & $1,200$ \\
Chunk Overlap Token Size & $100$ & $100$ \\
Community Report Max Length & $2,000$ & $-$ \\
Max Input Length & $8,000$ & $-$ \\
Max Cluster Size & $10$ & $-$ \\
Entity Summary Max Tokens & $-$ & $500$ \\
\bottomrule
\end{tabular}
% }
    \vskip -0.1in
\end{table}

% \begin{table}[h]
% \centering
% \caption{Parameters: LightRAG}
%     \vskip 0.1in
% \begin{tabular}{cc}
% \hline
% \textbf{Parameter} & \textbf{Value} \\
% \hline
% Mode & Local \\
% Response Type & Short phrase \\
% Top-k & 60 \\
% Chunk Token Size & 1200 \\
% Chunk Overlap Token Size & 100 \\
% Community Report Max Length & 2000 \\
% Max Input Length & 8000 \\
% Max Cluster Size & 10 \\
% \hline
% \end{tabular}
%     \vskip -0.1in
% \end{table}

% \begin{table}[h]
% \centering
% \caption{Parameters: GraphRAG}
%     \vskip 0.1in
% \begin{tabular}{cc}
% \hline
% \textbf{Parameter} & \textbf{Value} \\
% \hline
% Mode & Local \\
% Response Type & Short phrase \\
% Top-k & 60 \\
% Chunk Token Size & 1200 \\
% Chunk Overlap Token Size & 100 \\
% Entity Summary to Max Tokens & 500 \\
% \hline
% \end{tabular}
%     \vskip -0.1in
% \end{table}

\clearpage

%%%%%%%%%%%%%%%%%%%%%%%%%%%%%%%%%%%%%%%%%%%%%%%%%%%%%%%%%%%%%%%%%%%%%%%%%%%%%%%
%%%%%%%%%%%%%%%%%%%%%%%%%%%%%%%%%%%%%%%%%%%%%%%%%%%%%%%%%%%%%%%%%%%%%%%%%%%%%%%

\end{document}